%% file: aaai2027.tex
\documentclass[letterpaper]{article} 
\usepackage{aaai2027}  
\usepackage[hyphens]{url}  
\usepackage{graphicx} 
\usepackage{natbib}  
\usepackage{caption} 
\usepackage{algorithm}
\usepackage{algorithm}
\usepackage{algorithmic}
\usepackage{booktabs}
\usepackage{xcolor}
\usepackage{multirow}
\usepackage{tabularx}
\newcolumntype{Y}{>{\centering\arraybackslash}X}
\usepackage{array}
\usepackage{subcaption}
\usepackage{xcolor}
\usepackage{amsmath}
\usepackage{graphicx}
\usepackage{booktabs}

\usepackage[table]{xcolor}
\usepackage{tabularx}
\usepackage{array}
\usepackage{booktabs}

\usepackage[most]{tcolorbox}
\usepackage{listings}
\usepackage{amsmath}
\usepackage{xcolor}
\usepackage{newfloat}
\usepackage{listings}
\DeclareCaptionStyle{ruled}{labelfont=normalfont,labelsep=colon,strut=off} 
\floatstyle{ruled}
\newfloat{listing}{tb}{lst}{}
\floatname{listing}{Listing}
\usepackage{pgfplots}
\pgfplotsset{compat=1.18}

\title{SkillProx: Self-Evolving Agent Skills via Proximal Textual Gradient Descent}
\author {
    Mingxuan Zheng\textsuperscript{\rm 1}\equalcontrib,
    Yujin Zhou\textsuperscript{\rm 1}\equalcontrib,
    Chuxue Cao\textsuperscript{\rm 1}\equalcontrib,
    Boqin Yin\textsuperscript{\rm 1},\\
    Yuyao Zhang\textsuperscript{\rm 1},
    Jiapeng Sun\textsuperscript{\rm 1},
    Shuaishuai Gong\textsuperscript{\rm 2},
    Sirui Han\textsuperscript{\rm 1}\corresponding
    Yike Guo\textsuperscript{\rm 1}\corresponding
}
\affiliations {
    \textsuperscript{\rm 1}Hong Kong University of Science and Technology\\
    \textsuperscript{\rm 2}Macau University\\
    mzhengan@connect.ust.hk
}

\nocopyright
\begin{document}

\maketitle

\begin{abstract}

LLM agents increasingly adapt to recurring tasks by accumulating procedural knowledge in skills. These skills are lightweight, reusable textual artifacts that are loaded into the agent's context without weight updates. Recent methods refine skills through iterative task execution, failure diagnosis, and trajectory-guided text-space updates. However, existing frameworks lack explicit diagnosis--outcome feedback and treat deletion as a generic edit operation rather than a dedicated mechanism for consolidating accumulated knowledge. We introduce \textsc{SkillProx}, a proximal-gradient-inspired forward--backward framework that couples closed-loop diagnostic evolution with utility-aware proximal refinement. Motivated by a composite objective balancing task loss and skill complexity, the forward stage re-executes diagnosis-driven edits on the same task batch, rolls back regressions, and feeds measured outcomes into subsequent diagnoses. The backward stage decomposes the resulting skill into auditable knowledge units, estimates their contributions using a frozen leave-one-out utility audit, and applies validation-gated consolidation, demotion, or removal. Experiments on in-distribution and out-of-distribution benchmarks across multiple backbone LLMs show that \textsc{SkillProx} improves average accuracy by 3.0 percentage points over the strongest gradient-based baseline. Component ablations demonstrate the complementary effects of closed-loop diagnosis and proximal refinement. \footnote{Code will be available at \url{https://github.com/Steven011018/SkillProx}.}
\end{abstract}

\section{Introduction}

\input{sections/introduction}

\section{Related Works}

\input{sections/related_works}

\section{Preliminaries \& Motivation}
\label{sec:preliminaries}

\subsection{Problem Formulation}
\label{subsec:problem-formulation}
Let $X$ denote a complete structured skill artifact. A skill consists of a primary instruction file, \texttt{SKILL.md}, and an optional \texttt{references/} directory containing auxiliary instructions, examples, or resources. Let $\mathcal{X}$ denote the feasible space of structurally valid skill artifacts.
For any task set $D$, let $H_D(X)$ and $C_D(X)$ denote the hard accuracy and mean cell accuracy obtained using skill $X$, respectively. Let $G(X)$ denote the textual complexity of $X$, implemented as the total number of characters in \texttt{SKILL.md} and all active reference files.
Because a skill is structured, it can be decomposed into a collection of knowledge units
\begin{equation}
    \mathcal{Q}(X)
    =
    \{q_1,q_2,\ldots,q_m\},
\end{equation}
where each unit may correspond to an instruction section or a referenced resource. The concrete unit decomposition used by \textsc{SkillProx} is introduced in Section~\ref{subsec:skillprox-audit}.
Conceptually, task loss and textual complexity define the following composite optimization problem:
\begin{equation}
    \min_{X\in\mathcal{X}}
    \quad
    J_{\lambda}(X)
    :=
    L_{\mathcal{T}}(X)
    +
    \lambda G(X),
    \qquad
    \lambda\geq0,
    \label{eq:skillprox-objective}
\end{equation}
where $L_{\mathcal{T}}(X)$ denotes the expected loss of $X$ over an unknown task distribution $\mathcal{T}$, and $\lambda$ represents a conceptual trade-off between task performance and textual complexity.
Equation~\eqref{eq:skillprox-objective} characterizes the two objectives involved in skill evolution rather than an objective directly optimized by our implementation. Textual skills are discrete and non-differentiable, and our method does not take $\lambda$ as an explicit input. The formulation instead provides a common basis for examining performance-oriented skill updates and complexity-oriented knowledge consolidation, which we motivate empirically in the following two subsections.

\subsection{Motivation}
\textbf{Learning from Realized Update Effects.}
Across ten Qwen3.6-27B training runs, open-loop skill evolution achieves an average OJ hard accuracy of $50.30 \pm 2.50$, whereas a feedback-controlled variant achieves $51.40 \pm 1.51$.
Inspection of representative training trajectories shows that some updates appear reasonable from the diagnosis alone but reduce task performance after being incorporated into the skill. This discrepancy indicates that the quality of a textual update cannot be reliably determined from its semantic plausibility alone and instead needs to be assessed through its realized execution outcomes.
Re-evaluating the updated skill on the same task batch provides direct evidence of whether an update improves, preserves, or degrades performance. This evidence can be used both to determine whether the update should be retained and to inform subsequent diagnoses about previously effective or ineffective directions.
These observations motivate a closed-loop process of diagnosis, evaluation, and outcome feedback, in which the skill and its diagnostic process co-evolve according to the realized effects of proposed updates.


\textbf{Consolidating Accumulated Knowledge.}
Representative evolved skills exhibit repeated procedural instructions and task-specific solutions presented as reusable rules. Additional skill content therefore does not necessarily translate into additional task-solving capability.
Leave-one-out evaluation further identifies accumulated knowledge units with negative utility, meaning that their removal improves validation performance.
In one representative case, utility-aware consolidation reduces the skill size by 3.12\% while improving OJ hard accuracy from 46\% to 54\%.
These observations motivate a backward process that reassesses accumulated knowledge at the unit level and selectively removes or consolidates harmful content while preserving knowledge that contributes to task performance.
Case studies of the above description are provided in the Appendix~\ref{app:case_mov}.

\subsection{Proximal Gradient Descent}
\label{subsec:pgd-background}
The preceding observations motivate two complementary operations: a forward process that evaluates and improves newly proposed knowledge, and a backward process that reassesses and consolidates accumulated knowledge~\cite{tanabe2019proximal}. Proximal gradient descent provides a natural conceptual framework for relating these operations.
For a continuous composite objective
\begin{equation}
    \min_x
    \quad
    f(x)+\lambda g(x),
\end{equation}
proximal gradient descent first performs a forward gradient step on the differentiable objective $f$ and then applies a backward proximal step associated with the potentially non-smooth regularizer $g$:
\begin{equation}
    \begin{aligned}
        v_k
        &=
        x_k-\eta\nabla f(x_k),\\
        x_{k+1}
        &=
        \operatorname{prox}_{\eta\lambda g}(v_k),\\
        \operatorname{prox}_{\eta\lambda g}(v)
        &=
        \arg\min_x
        \left\{
            \lambda g(x)
            +
            \frac{\lVert x-v\rVert_2^2}{2\eta}
        \right\},
    \end{aligned}
    \label{eq:skillprox-pgd}
\end{equation}
where $\eta>0$ denotes the forward step size and $\lambda\geq0$ controls the regularization strength. The forward step primarily reduces the task objective, whereas the proximal step controls the structure or complexity of the resulting solution.
Skill text is discrete and non-differentiable and therefore does not admit a direct numerical gradient or a standard proximal operator. Our implementation also does not explicitly optimize Equation~\eqref{eq:skillprox-objective} or take $\eta$ and $\lambda$ as inputs.
We retain only the forward--backward division of responsibilities: diagnostic evolution serves as an inexact, performance-oriented forward operator, while utility-aware consolidation serves as a discrete backward operator for controlling accumulated knowledge. Section~\ref{sec:skillprox} instantiates this decomposition through closed-loop diagnostic evolution and validation-gated proximal refinement.

\input{tables/alignment_pgd}

\section{Method}
\label{sec:skillprox}

\subsection{Overview}
\label{subsec:skillprox-overview}

Figure~\ref{fig:method} summarizes the \textsc{SkillProx} pipeline. Following the forward--backward view introduced in Section~\ref{subsec:pgd-background}, \textsc{SkillProx} maps an initial skill $X_0$ to an intermediate skill $X_f$ through forward diagnostic evolution and subsequently produces the final skill $X^\star$ through backward proximal refinement. Forward iteration $k$ uses a training batch $B_k$, whereas utility auditing and backward validation use a fixed validation split $V$.
Together, the two stages realize \emph{diagnostic--proximal co-evolution}. Forward diagnosis evolves the task-solving content of the skill according to realized execution outcomes, while backward Prox evolves its structure by selectively preserving and consolidating accumulated knowledge.
The remainder of this section follows the execution order of the method. We first introduce the closed-loop forward update, followed by the frozen utility audit and candidate construction, validation-gated proximal refinement, and the implementation-level properties of the resulting procedure.


\subsection{Closed-Loop Forward Update}
\label{subsec:skillprox-forward}

At forward iteration $k$, the current skill $X_k$ is executed on training batch $B_k$. A diagnostician uses failed trajectories, contrastive successful trajectories, recent history, and the rejection reason from the previous attempt to propose an edit direction. The Patcher then produces candidate $\widetilde{X}_k^{(j)}$ for attempt $j$ from the same pre-iteration snapshot. The candidate is re-executed on the same batch and accepted according to
\begin{equation}
    \begin{aligned}
        \operatorname{Gate}_{\mathrm{fwd}}
        \!\left(
            \widetilde{X}_k^{(j)}
        \right)
        &=
        \mathbf{1}
        \!\left[
            H_{B_k}
            \!\left(
                \widetilde{X}_k^{(j)}
            \right)
            \geq
            H_{B_k}(X_k)
        \right] \\
        &\quad\times
        \mathbf{1}
        \!\left[
            C_{B_k}
            \!\left(
                \widetilde{X}_k^{(j)}
            \right)
            \geq
            C_{B_k}(X_k)
        \right].
    \end{aligned}
    \label{eq:skillprox-forward-gate}
\end{equation}

Here, $\mathbf{1}[\cdot]$ denotes the indicator function, which equals $1$ when its enclosed condition is true and $0$ otherwise. Consequently, the forward gate equals $1$ only when both the hard accuracy and mean cell accuracy satisfy their respective acceptance conditions.
The first candidate with a strict hard-accuracy gain terminates the search early. If no candidate strictly improves hard accuracy, the method evaluates at most three attempts and selects the lexicographically best attempted candidate by hard and cell accuracy. The winner must still satisfy Eq.~\eqref{eq:skillprox-forward-gate}; otherwise, the iteration keeps $X_k$ unchanged. For a rejected attempt, the observed hard/cell changes and attempted directions are provided to the next diagnosis. Across iterations, compact accept/reject summaries form a semantic history for subsequent diagnoses.

Equation~\eqref{eq:skillprox-forward-gate} constrains only the current training batch $B_k$. Because different iterations use different batches, it does not imply monotonic performance across iterations, on the validation split, or on the test set.

\subsection{Frozen Utility Audit and Candidate Selection}
\label{subsec:skillprox-audit}

After forward optimization, $X_f$ is parsed into $n$ auditable L2 sections and L3 reference groups:
\begin{equation}
\small
    \mathcal{Q}(X_f)
    =
    \left\{
        q_1,q_2,\ldots,q_n
    \right\}.
    \label{eq:skillprox-audit-units}
\end{equation}

Let $\operatorname{Ablate}(X_f,q_i)$ denote a copy of $X_f$ with unit $q_i$ fully removed. For an L2 unit, the corresponding section is removed. For an L3 unit, the reference file and all of its pointers are removed. We define the hard and cell marginal utilities of $q_i$ as
\begin{equation}
\small
    \begin{aligned}
        u_i^{\mathrm{hard}}
        &=
        H_V(X_f)
        -
        H_V
        \!\left(
            \operatorname{Ablate}(X_f,q_i)
        \right), \\
        u_i^{\mathrm{cell}}
        &=
        C_V(X_f)
        -
        C_V
        \!\left(
            \operatorname{Ablate}(X_f,q_i)
        \right).
    \end{aligned}
    \label{eq:skillprox-utility}
\end{equation}

A positive value in Eq.~\eqref{eq:skillprox-utility} means that removing the unit lowers performance; a negative value means that the ablated version performs better. All utilities are measured once before Prox and remain frozen throughout the candidate traversal.
Prox selects candidates directly from raw cell utility. The candidate set and processing order are
\begin{equation}
\small
    \begin{aligned}
        &\mathcal{I}_{\tau}
        =
        \left\{
            i\in\{1,\ldots,n\}:
            u_i^{\mathrm{cell}}<\tau
        \right\}, 
        \tau
        =
        -0.001, \\
        &i \prec j
        \Longleftrightarrow
        \left(
            u_i^{\mathrm{cell}},
            u_i^{\mathrm{hard}}
        \right) 
        <_{\mathrm{lex}}
        \left(
            u_j^{\mathrm{cell}},
            u_j^{\mathrm{hard}}
        \right).
    \end{aligned}
    \label{eq:skillprox-candidates}
\end{equation}
Here, $<_{\mathrm{lex}}$ denotes ascending lexicographic order, rather than an independent mathematical operator. Specifically, $(a,b)<_{\mathrm{lex}}(c,d)$ holds if $a<c$, or if $a=c$ and $b<d$. Therefore, candidates are first ordered by cell utility, with smaller values processed earlier, and ties are broken by hard utility.
The audit determines only candidate eligibility and processing order. It evaluates full ablation, whereas the Shrinker may consolidate, demote, or remove the target. Every realized trial must therefore be evaluated again in the current state.

\subsection{Prox: Single-Pass Validation-Gated Shrinkage}
\label{subsec:skillprox-prox}

Let $(i_1,\ldots,i_M)$ be the candidates ordered by Eq.~\eqref{eq:skillprox-candidates}, and initialize $X^{(0)}=X_f$. Each candidate is processed at most once. If an earlier consolidation has already removed its target, the candidate is skipped. Otherwise, the Shrinker produces a trial $T_m$ in a temporary copy of the current skill. The trial must first satisfy structural validity and strict complexity reduction:
\begin{equation}
\small
    \begin{aligned}
        \operatorname{StructOK}(T_m)&=1, 
        G(T_m)<G\!\left(X^{(m)}\right).
    \end{aligned}
    \label{eq:skillprox-structure-gate}
\end{equation}

If Eq.~\eqref{eq:skillprox-structure-gate} holds, $T_m$ is evaluated on the fixed validation split $V$. It must satisfy the hard, cell, and compression-range conditions
\begin{equation}
\small
    \begin{aligned}
        &H_V(T_m)
        \geq
        H_V\!\left(X^{(m)}\right)-\delta_h, \\
        &C_V(T_m)
        \geq
        C_V\!\left(X^{(m)}\right)-\delta_c, \\
        &1-
        \frac{
            G\!\left(X^{(m)}\right)
        }{
            G(X_f)
        }
        <
        \rho,
        \\
        &\delta_h=0,\delta_c=0.02, \rho=0.10.
    \end{aligned}
    \label{eq:skillprox-validation-gate}
\end{equation}
Here, $\delta_h\geq0$ and $\delta_c\geq0$ are the maximum absolute per-edit decreases allowed in hard accuracy and cell accuracy, respectively, while $\rho\in[0,1]$ is the cumulative compression threshold measured relative to the forward skill $X_f$. We set $\delta_h=0$, requiring recorded validation hard accuracy not to decrease, and $\delta_c=0.02$, allowing each accepted edit to reduce mean cell accuracy by at most $0.02$, or two percentage points when the metric is normalized to $[0,1]$. We set $\rho=0.10$.
This condition is evaluated on the active skill $X^{(m)}$ before the trial is committed, rather than on $T_m$ itself. Therefore, $\rho$ is not a strict upper bound on the compression ratio of the final output; it is a soft stopping threshold that determines whether another shrinkage attempt may proceed. As long as the current cumulative compression remains below $\rho$, the next trial may be accepted and move the final compression ratio beyond $\rho$.
The active state is updated as
\begin{equation}
\small
    \begin{aligned}
        X^{(m+1)}
        &=
        \begin{cases}
            T_m,
            & \text{if all gates pass}, \\
            X^{(m)},
            & \text{otherwise},
        \end{cases} \\
        X^\star
        &=
        X^{(M)}.
    \end{aligned}
    \label{eq:skillprox-state-update}
\end{equation}

All edits are first applied to a trial copy. A rejected trial is deleted, while an accepted trial replaces the active skill. A candidate that produces no accepted deletion does not terminate the loop; processing continues until the candidate sequence is exhausted. The compression condition in Eq.~\eqref{eq:skillprox-validation-gate} is evaluated on $X^{(m)}$ before applying $T_m$. Thus, $\rho=0.10$ is a soft cap: the final accepted edit may move the total compression beyond $10\%$.

\subsection{Implementation-Level Properties and Trade-offs}
\label{subsec:skillprox-properties-tradeoffs}

The Prox candidate set is finite, and each candidate is processed at most once; therefore, shrinkage terminates after a finite number of trials. 
Every accepted edit must strictly reduce text complexity, whereas a rejected trial leaves the active skill unchanged. 
The validation gate provides only a per-edit empirical constraint on recorded validation performance and does not imply test-set monotonicity or classical PGD convergence. 
Formal statements, cumulative performance bounds, and evaluation-complexity analysis are provided in Appendix~\ref{app:method_details}.

Shrinkage strength is jointly determined by the candidate threshold, performance tolerances, and compression range. 
Varying these controls produces skills with different text-compression ratios and held-out performance, yielding an empirical compression--performance trade-off. 
We study this relationship primarily through a sweep over the candidate threshold $\tau$; the sweep configuration, evaluation protocol, and Pareto filtering procedure are described in Subsection~\ref{pareto and skill size}.
Importantly, $\tau$ changes candidate coverage under the frozen utility audit but is not the explicit regularization weight $\lambda$ in the composite objective. 
The $\tau$-sweep should therefore be interpreted as a threshold-induced empirical trade-off curve rather than an exact $\lambda$-regularization path. 
The ideal regularized decision rule and its approximate relationship to $\tau$ are discussed in Appendix~\ref{app:method_details}.

\section{Experiments}

\input{sections/experiments}
\section{Conclusion}
In this paper, we present \textsc{SkillProx}, a proximal-gradient-inspired framework for self-evolving agent skills. \textsc{SkillProx} couples closed-loop diagnostic evolution, which verifies candidate updates through execution feedback, with utility-aware proximal refinement, which reassesses and consolidates accumulated knowledge. Across SpreadsheetBench Verified, WikiTableQuestions, and HiTab, \textsc{SkillProx} improves skill performance with multiple backbone LLMs under both in-distribution and out-of-distribution settings. Empirical results suggest a practical path for jointly improving diagnostic updates and accumulated skill knowledge through forward--backward optimization.
\newpage  
\bibliography{aaai2027}


\input{appendix}

\end{document}

%% file: sections/introduction.tex
\begin{figure*}[t]
    \centering
    \includegraphics[width=\textwidth]{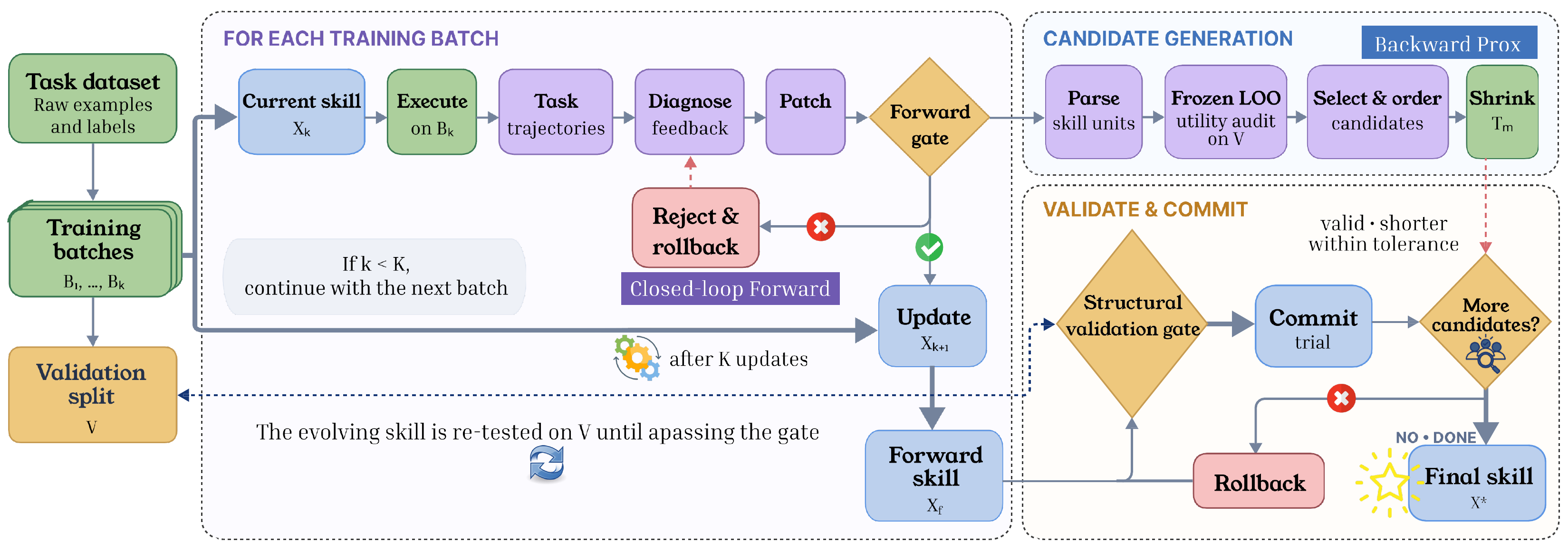}
\caption{\textbf{Pipeline of \textsc{SkillProx}.} The forward stage iteratively executes, diagnoses, and patches the current skill, using same-batch re-execution to accept beneficial updates or roll back rejected ones with outcome feedback. After $K$ updates, the backward stage audits the utility of knowledge units on a fixed validation split and applies validation-gated consolidation, demotion, or removal. Accepted edits produce the final skill $X^\star$.}
    \label{fig:method}
\end{figure*}

Large language model (LLM) agents have demonstrated strong capabilities in solving complex tasks through reasoning, tool use, and interaction with external environments~\cite{yao2023react, feng2025retoolreinforcementlearningstrategic, luo2025largelanguagemodelagent, tongyideepresearchteam2026tongyideepresearchtechnicalreport, kimiteam2026kimik25visualagentic}. To reuse procedural knowledge acquired from previous experience, recent agents increasingly externalize task-solving procedures, tool-use strategies~\cite{yuan2024easytoolenhancingllmbasedagents, lu2025toolsandboxstatefulconversationalinteractive, zhou2026lras, bai2026glance}, and domain-specific heuristics as persistent and editable \emph{skills}~\cite{xu2026agent, jiang2026sokagenticskills}. Previous studies primarily focus on skill synthesis, constructing reusable skills from demonstrations, successful trajectories, or external documents~\cite{chen2023skill}. However, benchmarks such as SkillsBench~\cite{li2026skillsbenchbenchmarkingagentskills} show that one-shot synthesis can provide uneven or even negative gains across heterogeneous tasks, since a skill constructed once cannot anticipate the diverse situation it will later encounter. This limitation motivates a shift from static skill construction to skill evolution. Representative methods such as SkillGrad~\cite{wang2026skillgrad} treat task failures as textual gradients, diagnosing failures as new tasks arrive and directly commiting the inference experience as a patch, thereby tuning a static skill artifact into an evolving form of procedural memory. 


Despite this progress, existing skill evolution methodologies face two critical limitations. 
(i) \textit{Unverified forward updates.} An LLM-generated diagnosis is commonly treated as a valid update direction without verifying the realized effect of the resulting skill edit. Once a patch is generated, it is typically committed directly, and its effectiveness is neither measured through re-execution nor provided as feedback to subsequent diagnoses. (ii) \textit{Unregulated skill growth.} Iterative patching continuously expands the skill without an explicit mechanism for reassessing accumulated knowledge. The resulting artifact may contain repetitive instructions, conflicting heuristics, or task-specific solutions incorrectly generalized as reusable rules, some of which can interfere with useful knowledge rather than merely increase textual complexity. 
This raises a central question: \textbf{\emph{how can outcome-verified forward diagnosis be coupled with structure-aware backward refinement, so that a skill's task capability and its accumulated structure co-evolve?}}

To explore this question, we conduct an analysis of open-loop, growth-oriented skill evolution, which points to two actionable design principles. On the forward side, we find that updates cannot be judged from their diagnostic text alone, but re-executing the updated skill on the same task batch directly reveals whether an update helps or hurts, suggesting that realized outcomes should gate whether an update is retained. On the backward side, we find that accumulated knowledge can be audited at the unit level, where removing negative-utility content improves accuracy from 46\% to 54\%, suggesting that skill health should be maintained through selective consolidation rather than continual growth. 

Guided by these principles, we propose \textbf{SkillProx}, a coupled forward--backward framework that realizes \emph{diagnostic--proximal co-evolution}. To resolve unverified forward updates, its forward component re-executes each candidate skill on the same task batch and commits the update only when it satisfies an outcome-grounded performance gate. Otherwise, the update is rolled back and retried, while its measured effect and rejected edit direction are retained as feedback for subsequent diagnoses. The diagnostic process therefore evolves according to the realized consequences of previous updates rather than their semantic plausibility alone. To counter unregulated skill growth, the backward component decomposes the accumulated skill into auditable knowledge units and estimates their contributions through a frozen leave-one-out utility audit. It then performs validation-gated proximal refinement, proposing consolidation, demotion, or removal edits and retaining only those that preserve structural validity and task performance. Together, the forward process determines which newly proposed knowledge should enter the skill, while the backward process determines which accumulated knowledge should persist, allowing diagnostic evolution and proximal consolidation to jointly shape the skill. Our main contributions are summarized as follows:

\begin{itemize}
    \item \textbf{A forward--backward formulation of skill evolution.}
    We formulate skill evolution as composite optimization over task performance and skill complexity, revealing two missing components in existing methods: outcome-grounded verification of forward updates and utility-guided regularization of accumulated knowledge.

    \item \textbf{The \textsc{SkillProx} framework.}
    We introduce \textsc{SkillProx}, which combines closed-loop diagnostic co-evolution including same-batch re-execution, rollback, retry, and accept/reject memory with validation-gated proximal shrinkage. This design enables online prevention of harmful edits and retrospective removal of negative-utility knowledge.

\item \textbf{Empirical validation of diagnostic--proximal co-evolution.}
Across three backbone LLMs, we compare \textsc{SkillProx} against six baselines on one in-distribution and two out-of-distribution benchmarks, observing performance gains in most evaluated settings. Component ablations further show that closed-loop diagnosis improves task accuracy, while proximal refinement provides additional gains through utility-aware knowledge consolidation.
    
   
\end{itemize}


%% file: sections/related_works.tex

Recent LLM agents increasingly externalize procedural knowledge into \emph{skills}—structured bundles of instructions, scripts, and resources that are loaded into the agent's context at inference time and improve task execution without any weight update~\cite{xia2026skillrlevolvingagentsrecursive,xu2026agentskillslargelanguage,xu2026agentskillslargelanguage,zhou2026externalizationllmagentsunified}.
Because authoring such skills by hand is label-intensive, hard to scale, and prone to human–machine cognitive misalignment~\cite{alzubi2026evoskillautomatedskilldiscovery,li2026skillsbenchbenchmarkingagentskills,yu2026skillonesizefitsallmodelawareskill,yang2026survey}, a first line of work studies \emph{skill synthesis}, which distills reusable skills from agent experience: Trace2Skill~\cite{ni2026trace2skilldistilltrajectorylocallessons} distills trajectories into skill artifacts, EvoSkill~\cite{alzubi2026evoskillautomatedskilldiscovery} discovers and edits skills through iterative failure analysis with validation-based selection, CoEvoSkills~\cite{zhang2026coevoskillsselfevolvingagentskills} couples a skill generator with a co-evolving surrogate verifier to build multi-file packages without ground-truth tests, and SkillComposer~\cite{zhang2026skillcomposerlearningevolveagent} build skills via decomposing skill construction into three learnable operations.

However, SkillsBench~\cite{li2026skillsbenchbenchmarkingagentskills} shows that one-shot synthesis is far from sufficient: curated skills help unevenly across domains, and self-generated skills yield negligible or even negative gains. Motivated by this observation, a second line of work builds on prior advances in text-space optimization, notably TextGrad's backpropagation of natural-language feedback~\cite{yuksekgonul2024textgradautomaticdifferentiationtext} and GEPA's reflective, Pareto-guided prompt evolution~\cite{agrawal2026gepareflectivepromptevolution}, and transplants these ideas to the skill artifact itself. SkillOpt~\cite{yang2026skilloptexecutivestrategyselfevolving} bounds each edit with a textual learning rate and a held-out gate, while SkillGrad~\cite{wang2026skillgradoptimizingagentskills} treats the skill package as a structured parameter updated by trajectory-level loss evidence, textual gradients, and momentum.
These methods, however, remain \emph{open-loop and growth-oriented}: a diagnosis is accepted as a valid update direction without verifying its realized effect, so its outcome is never fed back to subsequent diagnoses, and deletion is treated as one generic edit among many rather than a dedicated shrinkage mechanism, letting redundant, conflicting, or instance-specific content accumulate and interfere with useful knowledge.
SkillProx addresses both gaps by closing the diagnosis–validation loop in the forward step and introducing a utility-aware proximal step that explicitly optimizes the complexity term.

%% file: tables/alignment_pgd.tex
\begin{table*}[t]
    \centering
    \label{tab:skillprox-pgd-alignment}
    \small
    \begin{tabular}{p{0.16\linewidth}p{0.34\linewidth}p{0.40\linewidth}}
        \hline
        Component & Standard PGD & Proposed method \\
        \hline
        Objective
        & Composite task loss $f$ and complexity regularizer $g$
        & Two-stage task optimization and text complexity control \\
        \hline
        Forward direction
        & Analytical direction $-\nabla f$
        & Natural-language edit direction inferred from task trajectories \\
        \hline
        Forward update
        & $v_k=x_k-\eta\nabla f(x_k)$
        & Diagnose, patch, same-batch re-execute, reject feedback, and rollback \\
        \hline
        Backward step
        & Solve a proximal subproblem
        & Frozen utility audit, semantic shrinkage, structural checks, and validation gating \\
        \hline
        Locality
        & Distance penalty $\lVert x-v\rVert_2^2/(2\eta)$
        & Local single-target edits and a soft compression cap $\rho$ \\
        \hline
        Shrinkage strength
        & Controlled by $\eta\lambda$
        & Candidate, quality, and range controlled separately by $\tau$, $\delta_h$, $\delta_c$, and $\rho$ \\
        \hline
        Analysis
        & Convergence under smoothness and related assumptions
        & Finite termination, empirical gating, and strict complexity reduction \\
        \hline
    \end{tabular}
    \caption{Structural alignment between standard PGD and the proposed method.}
\end{table*}

%% file: sections/experiments.tex
\input{tables/main_table}

\subsection{Experimental Setups}

\paragraph{Benchmarks}
To comprehensively assess the effectiveness of our approach, we conduct experiments under both in-distribution (IID) and out-of-distribution (OOD) settings. For the IID evaluation, we adopt \textsc{SpreadsheetBench Verified}~\citep{ma2024spreadsheetbenchchallengingrealworld}, a human-validated subset of SpreadsheetBench that is specifically curated to enable reliable automatic evaluation. For the OOD evaluation, we further evaluate our method on \textsc{WikiTableQuestions}~\citep{pasupat2015compositionalsemanticparsingsemistructured} and \textsc{HiTab}~\citep{cheng2022hitabhierarchicaltabledataset}, which allows us to examine the generalization ability of the learned skills beyond the training distribution. Detailed descriptions of the above datasets are provided in the Appendix~\ref{app:exp_details}.

\paragraph{Baselines}
We compare our method against a broad range of baselines that span several representative categories.
First, we consider two fundamental settings: \textit{No Skill}, where the model operates without any auxiliary skill, and \textit{Human Skill}, where a manually curated skill is provided. Note that \textit{Human Skill} also serves as the base skill that SkillProx optimizes.
Second, for methods that leverage LLM-generated skills, we compare against \textsc{EvoSkill}~\citep{alzubi2026evoskillautomatedskilldiscovery} and \textsc{Trace2Skill}~\citep{ni2026trace2skilldistilltrajectorylocallessons}.
Third, for methods that focus on skill self-evolution, we compare against \textsc{SkillGrad}~\citep{wang2026skillgradoptimizingagentskills} and \textsc{SkillOpt}~\citep{yang2026skilloptexecutivestrategyselfevolving}.
All baselines and our method are evaluated on Qwen3.5-4B, Qwen3.5-27B, and Qwen3.6-27B to ensure a comprehensive and fair comparison across model scales and versions.

\paragraph{Implementation Details}
For the training configuration, we follow the settings of \textsc{SkillOpt} and \textsc{SkillGrad}. Specifically, we partition each dataset into training, validation, and test splits with a ratio of 2:1:8. The maximum number of interaction turns is set to 30 for all methods. For \textsc{EvoSkill} and \textsc{Trace2Skill}, we adopt their official configurations to guarantee a faithful comparison. For the evaluation protocol, we report accuracy as the primary metric. More details are provided in the Appendix~\ref{app:exp_details}.

\input{tables/ablation_table}

\subsection{Main results}

\paragraph{SkillProx reliably improves the base skill into a net-positive signal across all backbones.}
Starting from the Human Skill initialization it optimizes, SkillProx delivers consistent gains on every backbone: it lifts the base skill by $13.0$ pp on Qwen3.5-27B ($38.3 \rightarrow 51.3$) and by $17.8$ pp on Qwen3.6-27B ($36.7 \rightarrow 54.5$), while remaining slightly positive on the smaller Qwen3.5-4B ($20.3 \rightarrow 21.0$).
By treating the skill as an optimizable artifact rather than a fixed input, SkillProx converts the base initialization into stable, positive guidance.

\paragraph{SkillProx outperforms both LLM-generated and self-evolving skill baselines on the in-domain task.}
On SpreadsheetBench, SkillProx attains the best IID accuracy on all three backbones ($21.0$, $51.3$, and $54.5$).
The margin over LLM-generated skills is largest on the weaker 4B executor, where SkillProx exceeds EvoSkill by $14.3$ pp and Trace2Skill by $11.0$ pp, and it still edges out the strongest self-evolving baseline SkillGrad ($19.3 \rightarrow 21.0$).
It also exhibits the lowest variance among the self-evolving methods (e.g.\ $\pm0.5$ vs.\ $\pm7.6$ for SkillOpt on Qwen3.6-27B), indicating that its gains are stable across seeds rather than an artifact of favorable initialization.
Together, these results show that optimizing a skill yields more reliable behavior than merely generating one.

\paragraph{SkillProx generalizes robustly to out-of-domain tasks without overfitting to the training distribution.}
Although the skills are optimized solely on SpreadsheetBench, they transfer to OOD benchmarks with different formats and output spaces.
On WikiTQ, SkillProx achieves the best accuracy on Qwen3.5-4B ($78.5$), improving over No Skill by $13.5$ pp and over SkillGrad by $8.8$ pp, and is also best on Qwen3.5-27B ($86.8$) while remaining competitive on Qwen3.6-27B ($86.2$).
On HiTab it is best on both Qwen3.5-4B ($69.2$) and Qwen3.6-27B ($80.0$).
This robustness stands in sharp contrast to SkillOpt, which overfits to the in-domain style and collapses OOD, e.g.\ on the 4B model ($26.0$ on WikiTQ, $16.0$ on HiTab) and even on Qwen3.5-27B ($77.1$ and $66.8$).

\subsection{Ablation Studies}

We ablate the two core stages of SkillProx in Table ~\eqref{tab:ablation_spreadsheet}: the closed-loop diagnostic forward
update and the proximal shrinkage
backward stage, i.e.\ Prox. This yields
two variants. The first removes the closed-loop diagnosis and keeps only
proximal gradient descent. The second keeps the closed-loop forward
update but removes Prox. All component
ablations use Qwen3.6-27B with the same training configuration and a fixed
training-set seed. We adopt the same evaluation protocol and metric as the main
experiments, reporting results on the in-domain SpreadsheetBench setting.

Table~\ref{tab:ablation_spreadsheet} shows that removing either stage degrades
held-out accuracy relative to the full method. Removing the closed-loop
diagnosis lowers accuracy from $54.5$ to $53.0$ ($-1.5$ pp), while removing
Prox lowers it further to $52.0$ ($-2.5$ pp). The larger drop from removing
Prox indicates that task-driven forward editing alone accumulates redundant or
overly instance-specific content, which the proximal shrinkage stage is needed
to control; the smaller but consistent drop from removing the closed-loop
diagnosis indicates that Prox still requires a well-optimized forward skill to
shrink, rather than the raw base skill. The full method also attains the lowest
variance ($\pm0.5$ vs.\ $\pm1.0$), suggesting that the two stages act
complementarily to produce stable gains rather than an artifact of a favorable
initialization.

\subsection{Detailed Analysis}
\label{pareto and skill size}
\begin{figure}[t]
    \centering
    \includegraphics[width=1.0\linewidth]
    {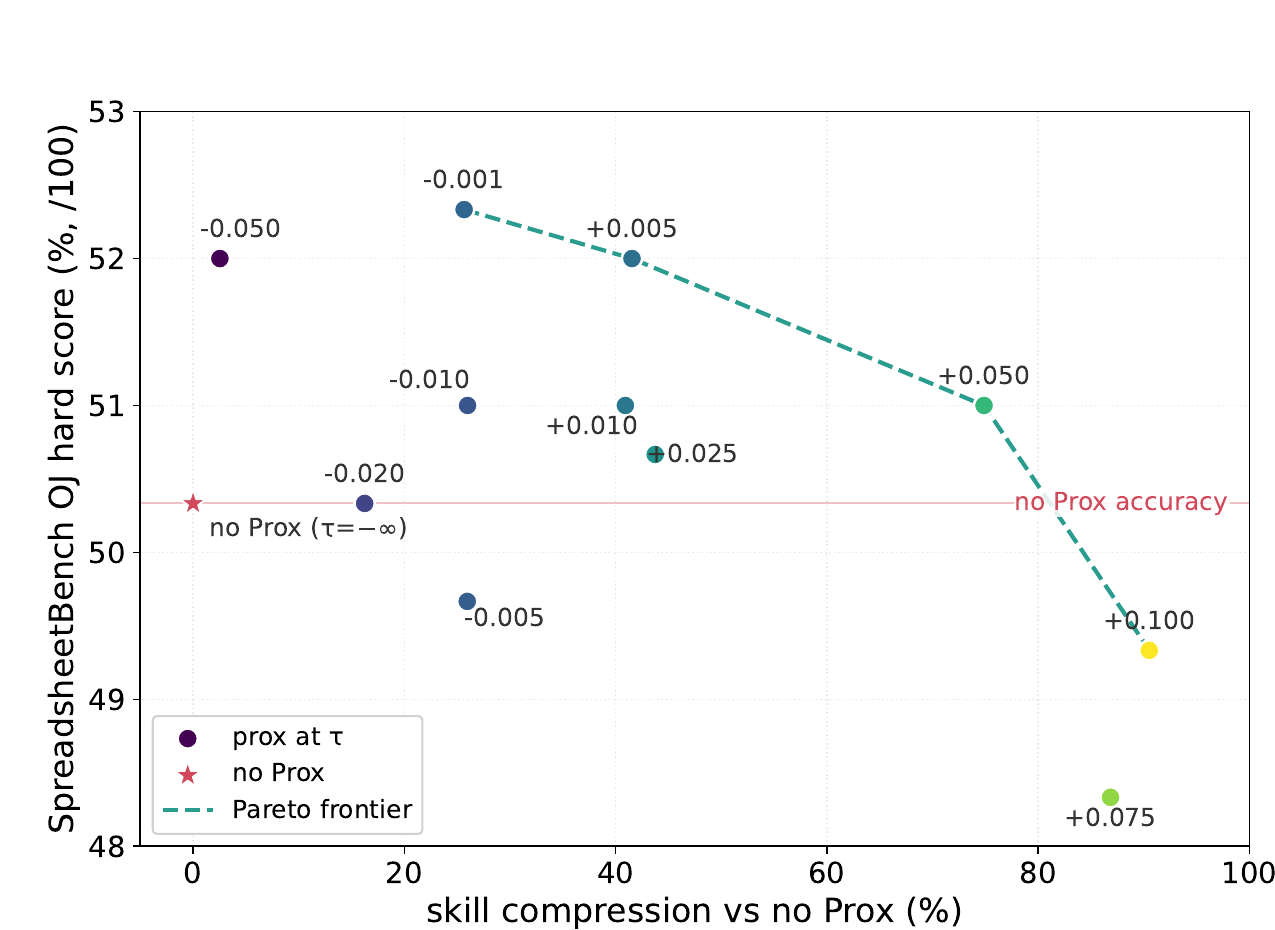}
    \caption{Accuracy--compression Pareto frontier of Qwen3.6-27B under
    different proximal thresholds $\tau$.}
    \label{fig:tau-pareto-en}
\end{figure}

\textbf{Accuracy--Compression Analysis over the Proximal Threshold $\tau$.}
In Figure ~\eqref{fig:tau-pareto-en}, we apply Prox offline to three closed-loop skills produced by Qwen3.6-27B without retraining. We evaluate the no-Prox anchor ($\tau=-\infty$), five negative thresholds, and six positive thresholds. 
The no-Prox setting achieves an OJ hard accuracy of $50.3\%$.
At $\tau=-0.001$, accuracy reaches its maximum of $52.3\%$ with
$25.7\%$ compression. At $\tau=0.005$, the compression ratio increases to
$41.5\%$ while accuracy remains at $52.0\%$. Even at $\tau=0.050$,
$74.9\%$ of the skill is removed while retaining $51.0\%$ accuracy.
Performance begins to decline when compression exceeds approximately $80\%$. Several Prox settings achieve both higher compression and higher accuracy than no Prox, indicating that
moderate shrinkage may also reduce duplicated rules and context interference.

\textbf{Model Size and Final Skill Length.} We compare the final G3D skills produced by Qwen3.5-4B and Qwen3.5-27B after closed-loop training and proximal shrinkage. 
\begin{table}[t]
    \centering
    \small
    \setlength{\tabcolsep}{5pt}
    \begin{tabular}{lcc}
        \toprule
        Metric & Qwen3.5-4B & Qwen3.5-27B \\
        \midrule
        Total length
        & $40.4\mathrm{k}\pm8.6\mathrm{k}$
        & $27.9\mathrm{k}\pm12.9\mathrm{k}$ \\
        Main file
        & $14.8\mathrm{k}\pm3.3\mathrm{k}$
        & $15.4\mathrm{k}\pm3.5\mathrm{k}$ \\
        References
        & $25.6\mathrm{k}\pm9.1\mathrm{k}$
        & $12.6\mathrm{k}\pm11.1\mathrm{k}$ \\
        Prox compression
        & $29.4\%$
        & $19.0\%$ \\
        \bottomrule
    \end{tabular}
\caption{Final skill length across model sizes. Results are averaged over three 4B runs and 27B runs.}
\label{tab:model-size-skill-length}
\end{table}
The final 4B skill is approximately $45\%$ longer than the 27B skill on average. Interestingly, the main skill files converge to nearly the same length of approximately $15\mathrm{k}$ characters. The difference arises almost entirely from reference files: the 4B model produces roughly twice as much reference content as the 27B model. This suggests that model size affects the amount of supplementary guidance required rather than the size of the core skill structure. 
Together, these results suggest that larger models naturally produce more selective and compact skills, while closed-loop validation and proximal shrinkage provide smaller models with an external mechanism for rejecting unnecessary updates. Consistent with this interpretation, G3D length is negatively correlated with IID hard accuracy within the 27B runs, indicating that a longer skill is not necessarily a more effective one. Further results are provided in the Appendix~\ref{app:add_results}.

%% file: tables/main_table.tex
\begin{table*}[ht]
\centering
\renewcommand{\arraystretch}{0.9}
\begin{tabularx}{\textwidth}{ll *{3}{Y}}
\toprule
\multirow{2}{*}{Model} & \multirow{2}{*}{Method}
 & \multicolumn{1}{c}{IID} & \multicolumn{2}{c}{OOD} \\
\cmidrule(lr){3-3} \cmidrule(lr){4-5}
 & & Spreadsheet & WikiTQ & HiTab \\
\midrule
\multirow{7}{*}{Qwen3.5-4B}
 & No Skill          & \underline{$20.3_{\pm1.5}$} & $65.0_{\pm3.3}$ & $61.7_{\pm2.0}$ \\
 & Human Skill       & \underline{$20.3_{\pm4.2}$} & $68.3_{\pm1.6}$ & $63.5_{\pm0.9}$ \\
 & EvoSkill          & $6.7_{\pm2.1}$ & \underline{$76.8_{\pm8.1}$} & $63.3_{\pm2.3}$ \\
 & Trace2Skill       & $10.0_{\pm0.0}$ & $66.0_{\pm2.3}$ & $57.5_{\pm1.7}$ \\
 & SkillOpt          & $15.3_{\pm2.5}$ & $26.0_{\pm6.6}$ & $16.0_{\pm1.3}$ \\
 & SkillGrad         & $19.3_{\pm0.5}$ & $69.7_{\pm0.8}$ & \underline{$65.4_{\pm3.4}$} \\
 & SkillProx (ours)  & $\mathbf{21.0}_{\pm1.4}$ & $\mathbf{78.5}_{\pm2.9}$ & $\mathbf{69.2}_{\pm2.3}$ \\
\midrule
\multirow{7}{*}{Qwen3.5-27B}
 & No Skill          & $44.0_{\pm6.1}$ & $85.3_{\pm1.8}$ & \underline{$78.5_{\pm1.0}$} \\
 & Human Skill       & $38.3_{\pm7.6}$ & $85.5_{\pm0.0}$ & $\mathbf{78.7}_{\pm1.3}$ \\
 & EvoSkill          & $8.7_{\pm3.5}$ & \underline{$86.0_{\pm0.5}$} & $\mathbf{78.7}_{\pm0.3}$ \\
 & Trace2Skill       & $32.0_{\pm9.6}$ & $84.2_{\pm0.8}$ & $76.7_{\pm1.9}$ \\
 & SkillOpt          & $51.3_{\pm4.7}$ & $77.1_{\pm1.9}$ & $66.8_{\pm0.3}$ \\
 & SkillGrad         & \underline{$51.3_{\pm0.9}$} & $85.2_{\pm0.6}$ & $77.5_{\pm0.4}$ \\
 & SkillProx (ours)  & $\mathbf{51.3}_{\pm1.2}$ & $\mathbf{86.8}_{\pm1.1}$ & \underline{$78.5_{\pm1.1}$} \\
\midrule
\multirow{7}{*}{Qwen3.6-27B}
 & No Skill          & $45.3_{\pm4.7}$ & $85.8_{\pm0.8}$ & $78.2_{\pm1.0}$ \\
 & Human Skill       & $36.7_{\pm8.1}$ & $85.7_{\pm1.2}$ & $78.0_{\pm0.9}$ \\
 & EvoSkill          & $13.0_{\pm4.4}$ & $\mathbf{87.7}_{\pm1.8}$ & $77.8_{\pm2.0}$ \\
 & Trace2Skill       & $35.3_{\pm3.1}$ & $85.2_{\pm0.6}$ & $78.7_{\pm2.3}$ \\
 & SkillOpt          & \underline{$53.3_{\pm7.6}$} & $82.2_{\pm0.6}$ & \underline{$79.7_{\pm0.8}$} \\
 & SkillGrad         & $50.0_{\pm3.3}$ & $84.8_{\pm0.6}$ & $78.3_{\pm0.3}$ \\
 & SkillProx (ours)  & $\mathbf{54.5}_{\pm0.5}$ & \underline{$86.2_{\pm0.3}$} & $\mathbf{80.0}_{\pm1.3}$ \\
\bottomrule
\end{tabularx}
\caption{Main results across three benchmarks. R is the average score
(0--100, $\uparrow$); values are mean$\pm$std over seeds. Spreadsheet is the
in-domain setting; WikiTQ and HiTab are out-of-domain. Best in
\textbf{bold}, second-best \underline{underlined}.}
\label{tab:main_results}
\end{table*}

%% file: tables/ablation_table.tex
\begin{table}[ht]
\centering
\renewcommand{\arraystretch}{0.9}
\begin{tabularx}{\columnwidth}{X r r}
\toprule
Variant & Mean$\pm$Std & $\Delta$ Acc. \\
\midrule
w/o closed-loop diagnosis & $53.0_{\pm1.0}$ & $-1.5$ \\
w/o Prox                  & $52.0_{\pm1.0}$ & $-2.5$ \\
SkillProx (full)          & $\mathbf{54.5}_{\pm0.5}$ & $-$ \\
\bottomrule
\end{tabularx}
\caption{Component ablation on SpreadsheetBench with
Qwen3.6-27B. The variant ``w/o closed-loop diagnosis'' keeps only proximal
gradient descent (Prox); the variant
``w/o Prox'' keeps the closed-loop forward update; the full method combines both stages. Best in \textbf{bold}.}
\label{tab:ablation_spreadsheet}
\end{table}

%% file: appendix.tex
\newenvironment{apptable}[1][\linewidth]
  {\par\medskip\noindent\begin{minipage}{#1}\centering\small}
  {\end{minipage}\par\medskip}
\newenvironment{appfigure}[1][\linewidth]
  {\par\medskip\noindent\begin{minipage}{#1}\centering}
  {\end{minipage}\par\medskip}
\definecolor{CaseTableHeader}{RGB}{242,245,244}
\definecolor{CaseTableBorder}{RGB}{204,204,204}
\definecolor{CasePositive}{RGB}{39,174,96}
\definecolor{CaseNegative}{RGB}{192,57,43}
\definecolor{ProxBefore}{RGB}{184,196,204}
\definecolor{ProxAfter}{RGB}{74,157,104}

\onecolumn
\appendix
\setcounter{secnumdepth}{3} 
\begin{center}
{\LARGE\bfseries Appendix}
\end{center}
\vspace{0.5em}

\section{Case Study for Motivation}\label{app:case_mov}

\subsection{Case Study: Removing Negative-Utility Redundancy}
\label{app:motivation-case}
Our method addresses two distinct failure modes in iterative skill evolution. The forward stage must determine whether a proposed update actually improves behavior, while the backward stage must remove redundant or harmful content that remains after repeated accumulation. These two problems require different forms of feedback.

\subsubsection{Forward Motivation: Outcome-Blind Skill Updates}
\label{sec:motivation-forward}
Standard forward skill evolution follows an execute--diagnose--patch procedure but does not re-execute the patched skill before committing the update. Consequently, a plausible diagnosis may be converted into a permanent rule without evidence that it improves task outcomes.
We observe this failure in a Qwen3.6-27B run involving sequential scans with a dynamically changing reference value. Open-loop evolution converts the failure into the meta-instruction ``trace a concrete example before coding'' and includes a task-specific template with the hard-coded condition \texttt{value >= reference * 1.10}. Leave-out auditing assigns negative estimated utility to this section:
\begin{equation}
u^{\mathrm{cell}}_{\mathrm{open}}=-0.0337,
\qquad
u^{\mathrm{hard}}_{\mathrm{open}}=-0.0556.
\label{eq:open-loop-negative-utility}
\end{equation}
Closed-loop evolution encounters the same underlying failure pattern but re-executes each proposed update on the current batch. Regressive candidates are rolled back, and their observed hard/cell changes are returned to the next diagnostic attempt. The resulting skill instead learns an actionable decision rule: after each qualifying event, update the reference to the current value before continuing the scan. This formulation has positive estimated utility:
\begin{equation}
u^{\mathrm{cell}}_{\mathrm{closed}}=+0.0495,
\qquad
u^{\mathrm{hard}}_{\mathrm{closed}}=+0.0474.
\label{eq:closed-loop-positive-utility}
\end{equation}
The contrast motivates an outcome-gated forward process. Semantic plausibility alone cannot distinguish a reusable rule from an over-specific template; same-batch re-execution provides a measurable signal for accepting, rejecting, or revising an update.

\subsubsection{Backward Motivation: Residual Redundancy and Negative Utility}
\label{sec:motivation-backward}
Outcome gating does not eliminate the need for post-hoc refinement. It evaluates an update only on the batch that produced it and therefore cannot reliably detect content that is harmless locally but redundant or harmful across a broader validation distribution. Indeed, the closed-loop skill in the same run still contains two negative-utility and fourteen zero-utility units. Notably, the main \emph{Sequential Scan with Dynamic Reference} section is useful, while its supporting reference file is assigned negative utility. Useful and harmful content may therefore coexist within the same knowledge group.
Forward evolution also accumulates repeated formulations across iterations. In the open-loop skill, the instruction ``trace a concrete example before coding'' appears in four sections, with cell utilities ranging from $+0.1038$ to $-0.0337$. The lowest-utility occurrence contains the hard-coded task-specific template described above.
A frozen utility audit identifies five candidate units in this run, but only one candidate survives re-evaluation against the active skill. The accepted edit consolidates the transferable state-tracing principle into a positive-utility interpretation section while removing the repeated template. It reduces the complete skill from 29,129 to 28,219 characters, corresponding to
\begin{equation}
\rho
=
\frac{29{,}129-28{,}219}{29{,}129}
=
3.12\%.
\label{eq:motivation-compression}
\end{equation}
The edit preserves validation hard accuracy at $94.74\%$ and increases validation cell accuracy from $96.05\%$ to $99.73\%$. This motivates a validation-gated backward Prox stage: the frozen audit identifies and orders potentially redundant units, while sequential validation determines whether each realized edit can be safely committed.

\subsubsection{Complementary Roles}
\label{sec:motivation-complementarity}
The two stages address complementary timescales. Closed-loop forward evolution provides \emph{online interception}: it measures the immediate effect of a proposed update and blocks same-batch regressions. Backward Prox provides \emph{retrospective refinement}: it reassesses the accumulated skill on a frozen validation set and removes redundancy that was not detectable from the originating batch.
Their respective roles can be summarized as
\begin{equation}
\text{Closed-loop forward}
=
\text{online update verification},
\label{eq:forward-role}
\end{equation}
\begin{equation}
\text{Backward Prox}
=
\text{post-training utility refinement}.
\label{eq:backward-role}
\end{equation}
Closed-loop feedback improves how knowledge is introduced, whereas Prox controls what remains after accumulation. Full case studies, candidate-level validation decisions, and OJ results are provided in Appendix~\ref{app:prox-case-study} and Appendix~\ref{app:closed-loop-case-study}.

\section{Additional Method Details}\label{app:method_details}

\subsection{Algorithmic Properties and Complexity}
\label{subsec:skillprox-properties}

The candidate set is finite, and each candidate is processed at most once, so the main loop terminates. Every accepted trial must be strictly smaller than the active skill, while a rejected trial leaves the active state unchanged. Each accepted edit also preserves recorded validation hard accuracy up to $\delta_h=0$ and may reduce recorded cell accuracy by at most $\delta_c$.

If the method accepts $R$ shrinkage edits, the recorded validation metrics satisfy
\begin{equation}
    \begin{aligned}
        G(X^\star)
        &\leq
        G(X_f), \\
        G(X^\star)
        &<
        G(X_f)
        \quad \text{if } R>0, \\
        H_V(X^\star)
        &\geq
        H_V(X_f), \\
        C_V(X^\star)
        &\geq
        C_V(X_f)-R\delta_c .
    \end{aligned}
    \label{eq:skillprox-properties}
\end{equation}

The complete audit requires one baseline evaluation and $n$ leave-one-out evaluations. Prox makes at most $M$ Shrinker calls and $M$ trial evaluations. If each evaluation executes all $|V|$ validation tasks, the number of task executions is
\begin{equation}
    \mathcal{O}
    \!\left(
        (n+M)|V|
    \right),
    \qquad
    M\leq n.
    \label{eq:skillprox-complexity}
\end{equation}
Equation~\eqref{eq:skillprox-properties} concerns the recorded validation scores. Accounting for stochastic evaluation and adaptive reuse of the same validation split requires a confidence bound that holds jointly over all queries; such a statistical analysis can be placed in the appendix.

\subsection{Empirical Compression--Performance Trade-off}
\label{subsec:skillprox-tradeoff}

The implementation has no explicit $\lambda$, so we do not interpret its experimental curve as an exact $\lambda$-regularization path. Let $\theta=(\tau,\delta_h,\delta_c,\rho)$ denote a shrinkage configuration. Each configuration produces a compression--held-out-performance point
\begin{equation}
    \begin{aligned}
        &\operatorname{Comp}(X^\star_{\theta})
        =
        1-
        \frac{
            G(X^\star_{\theta})
        }{
            G(X_f)
        }, \\
        &p(\theta)
        =
        \left(
            \operatorname{Comp}(X^\star_{\theta}),
            A_{\mathrm{test}}(X^\star_{\theta})
        \right),
    \end{aligned}
    \label{eq:skillprox-tradeoff-point}
\end{equation}
where $A_{\mathrm{test}}$ is the final performance on a fixed held-out test or OJ set. Aggregating points across skill seeds and repeated inference runs, followed by non-dominance filtering, yields an empirical Pareto approximation.

To isolate how the candidate range affects compression and performance, we perform a $\tau$-sweep using the same forward skill and frozen utility audit:
\begin{equation}
    \tau
    \in
    \left\{
        -0.05,\,
        -0.001,\,
        0.025,\,
        0.10
    \right\}.
    \label{eq:skillprox-tau-grid}
\end{equation}

As $\tau$ increases, the frozen-utility candidate set expands monotonically. Negative thresholds include only clearly negative-utility units, whereas positive thresholds also admit zero-utility and weakly positive-utility units. Each threshold is evaluated independently on the same held-out OJ set, and the uncompressed $X_f$ serves as the zero-compression anchor. Numerical results, cross-seed variability, dominated points, and knee points are analyzed in the experiments.

The threshold $\tau$ is not identical to the regularization weight $\lambda$. Let $c_i>0$ denote the complexity reduction produced by candidate operation $i$. An ideal single-candidate decision under explicit regularization would satisfy
\begin{equation}
    u_i
    \leq
    \lambda c_i.
    \label{eq:skillprox-lambda-criterion}
\end{equation}

If candidate complexity reductions are approximately equal to a representative value $\bar{c}$, the raw-utility threshold $u_i\leq\tau$ admits the approximation
\begin{equation}
    \lambda_{\mathrm{eff}}
    \approx
    \frac{\tau}{\bar{c}}.
    \label{eq:skillprox-effective-lambda}
\end{equation}

In the general case, the raw-$\tau$ rule corresponds to candidate-specific implicit weights
\begin{equation}
    \lambda_i
    =
    \frac{\tau}{c_i}.
    \label{eq:skillprox-implicit-lambda}
\end{equation}

Thus, $\tau$ is a monotone proxy for regularization pressure rather than a shared Lagrange multiplier. The regime $\tau<0$ performs safety-margined negative-utility screening, while $\tau>0$ approximately permits performance--complexity exchange.

To isolate the effect of $\tau$ on candidate coverage, the current diagnostic sweep sets both hard and cell tolerances to $1.0$ and does not impose the $10\%$ compression cap. It therefore measures a threshold-induced compression--accuracy curve rather than the deployment behavior of full Prox or an explicit $\lambda$-regularization path.

\section{Experimental Details}
\label{app:exp_details}
This section records the complete setup for the experiments with Qwen3.5-4B, Qwen3.5-27B, and Qwen3.6-27B, including the ten-run Qwen3.6-27B ablation and the two case studies below. Unless otherwise stated, conditions paired within a backbone use the same data split, training-task selection rule, initial skill, and evaluation protocol.

\subsection{Compared Conditions and Backbones}
\label{app:experimental-conditions}
The four evolved-skill conditions form a two-by-two component design, as summarized in Table~\ref{tab:experimental-conditions}. This design isolates the effects of closing the forward loop and applying backward Prox refinement.

\begin{apptable}
    \captionof{table}{Component design of the evolved-skill conditions.}
    \label{tab:experimental-conditions}
    \begin{tabular}{lcc}
        \hline
        Forward evolution & Without Prox & With Prox \\
        \hline
        Open-loop   & G1  & G2D \\
        Closed-loop & G3f & G3D (\textsc{SkillProx}) \\
        \hline
    \end{tabular}
\end{apptable}

We additionally evaluate two external baselines: \textbf{no skill}, in which the executor receives only a minimal skill stub, and \textbf{human skill}, in which it receives a manually authored spreadsheet skill. The three backbones and their training-seed coverage are summarized in Table~\ref{tab:experimental-backbones}.

\begin{apptable}
    \captionof{table}{Backbone models and training-seed coverage in the recorded experiment artifacts.}
    \label{tab:experimental-backbones}
    \begin{tabular}{lcc}
        \hline
        Backbone & Training seeds & Number of seeds \\
        \hline
        Qwen3.5-4B  & 0--2 & 3 \\
        Qwen3.5-27B & 0--2 & 3 \\
        Qwen3.6-27B & 0--2 & 3 \\
        \hline
    \end{tabular}
\end{apptable}

\begin{apptable}
    \captionof{table}{Core training, refinement, and evaluation parameters.}
    \label{tab:experimental-parameters}
    \begin{tabular}{p{0.20\linewidth}p{0.28\linewidth}p{0.44\linewidth}}
        \hline
        Stage & Parameter & Setting \\
        \hline
        Generation
        & Temperature / thinking
        & 0.7 / disabled \\
        Forward evolution
        & Training tasks / batch size
        & Up to 40 / 4 \\
        Forward evolution
        & Planned updates / early stopping
        & 10 / four consecutive all-correct batches \\
        Agent runtime
        & Maximum turns
        & Executor 30; Diagnoser 15; Momentum 15; Patcher 20; Shrinker 20 \\
        Closed-loop
        & Attempts / hard tolerance / prior size
        & 3 / 0 / 6 records \\
        Closed-loop
        & Acceptance gate
        & Non-decreasing hard-correct count and mean cell accuracy \\
        Utility audit
        & Validation tasks / test cases
        & 20 / 1 \\
        Utility audit
        & $\epsilon_{\mathrm{cell}}$
        & 0.025 for most runs \\
        Prox
        & Candidate threshold
        & $\tau=-0.001$\\
        Prox
        & Hard tolerance / cell tolerance
        & $\delta_h=0$ / $\delta_c=0.02$ \\
        Prox
        & Compression cap
        & $\rho=0.10$ (soft cap) \\
        OJ evaluation
        & Tasks / test cases / recalculation timeout
        & 100 / 3 / 180 seconds \\
        OOD evaluation
        & Examples / test cases
        & 200 per benchmark / 1 \\
        \hline
    \end{tabular}
\end{apptable}

\subsection{Execution and Grading Infrastructure}
\label{app:experimental-infrastructure}
Model requests are dispatched to vLLM endpoints with round-robin selection and endpoint failover. Spreadsheet outputs are recalculated with LibreOffice before grading, and concurrent recalculation is serialized through a shared office lock. Evaluation tasks run in isolated working directories so that model-generated helper files do not modify the repository or other task instances. Model inference is served on NVIDIA H800 GPUs with 80\,GB of memory per GPU. More details are provided in Table~\ref{tab:experimental-parameters}.

\section{Additional Results and Analysis}\label{app:add_results}
This appendix presents two complementary case studies on Qwen3.6-27B.
The first examines closed-loop forward evolution as online interception of
regressive updates. The second examines backward Prox as retrospective removal
of residual redundancy. Both studies reuse the same training seed in order to
connect the two mechanisms; they should therefore not be treated as independent
statistical evidence.

Besides, we provide more details on the analysis of model size and skill length.

\subsection{Case Study: Why Closed-Loop Forward Evolution Helps}
\label{app:closed-loop-case-study}

\subsubsection{Ten-Seed Aggregate Comparison}
\label{app:closed-loop-seeds}

We compare open-loop G1 and closed-loop G3f across ten Qwen3.6-27B training
seeds. For each seed, the two conditions use exactly the same training IDs and
batch partition. Evaluation uses the canonical OJ protocol: 100 fixed test
tasks, three test cases per task, and hard success only when all three cases
are cell-perfect.

\begin{apptable}
    \captionof{table}{Canonical same-batch OJ results across ten Qwen3.6-27B training
    seeds. G1 and G3f use identical training IDs and batch partitions. Seed~8,
    which attains the largest gain in both metrics, is highlighted.}
    \label{tab:closed-loop-seeds}
    \setlength{\tabcolsep}{7pt}
    \begin{tabular}{rrrrrrr}
        \toprule
        Seed
        & G1 hard & G3f hard & $\Delta$ hard
        & G1 cell & G3f cell & $\Delta$ cell \\
        \midrule
        0 & 47 & 51 & $+4$ & 72.33 & 75.60 & $+3.27$ \\
        1 & 50 & 51 & $+1$ & 75.49 & 75.48 & $-0.01$ \\
        2 & 49 & 49 & $0$  & 78.40 & 74.91 & $-3.49$ \\
        3 & 52 & 53 & $+1$ & 77.36 & 80.26 & $+2.90$ \\
        4 & 50 & 52 & $+2$ & 74.56 & 78.65 & $+4.09$ \\
        5 & 53 & 53 & $0$  & 77.47 & 77.55 & $+0.08$ \\
        6 & 51 & 52 & $+1$ & 77.96 & 78.22 & $+0.26$ \\
        7 & 51 & 49 & $-2$ & 78.00 & 77.85 & $-0.15$ \\
        \rowcolor{green!7}
        \textbf{8} & \textbf{46} & \textbf{51} & $\mathbf{+5}$
                   & \textbf{74.71} & \textbf{79.69} & $\mathbf{+4.98}$ \\
        9 & 54 & 53 & $-1$ & 78.68 & 79.88 & $+1.20$ \\
        \bottomrule
    \end{tabular}
\end{apptable}

Across the ten seeds, closed-loop evolution records six wins, two ties, and two
losses in hard accuracy, with a mean hard improvement of 1.10 points
(and a mean cell improvement of 1.31 points;
Table~\ref{tab:closed-loop-seeds}). The cross-seed standard deviation decreases
from 2.50 to 1.51, and the minimum increases from 46 to 49. The largest gain
occurs at seed~8 ($46\%\rightarrow51\%$ hard; $74.71\%\rightarrow79.69\%$
cell). However, seed~8 is also the weakest open-loop run, and its closed-loop
result ($51\%$) is close to the G3f mean rather than an unusually strong
closed-loop outcome. The more defensible interpretation is therefore that
closed-loop evolution stabilizes the lower tail rather than uniformly raising
the upper bound.

As a descriptive summary, the correlation between open-loop hard accuracy and
the hard improvement is
\begin{equation}
\operatorname{corr}
\left(
H_{\mathrm{G1}},
H_{\mathrm{G3f}}-H_{\mathrm{G1}}
\right)
=
-0.800.
\label{eq:closed-loop-corr}
\end{equation}
This quantity is informative but should not be over-interpreted: because the
baseline appears in both arguments, the negative sign is partly a consequence
of mathematical coupling and regression to the mean.

\paragraph{No measurement after an open-loop edit.}
The released run artifacts show that an open-loop iteration follows
\textsc{Execute} $\rightarrow$ \textsc{Classify} $\rightarrow$
\textsc{Diagnose} $\rightarrow$ \textsc{Momentum} $\rightarrow$
\textsc{Patch}. It does not execute the updated skill again, and all ten
open-loop patches for seed~8 are committed without an acceptance test.
Closed-loop evolution instead re-executes the same four-task batch after each
patch and applies
\begin{equation}
\mathrm{accept}
=
\mathbf{1}\!\left[
H_{\mathrm{post}}\ge H_{\mathrm{pre}}-\tau_h
\ \land\
S_{\mathrm{post}}\ge S_{\mathrm{pre}}-10^{-9}
\right],
\qquad \tau_h=0,
\label{eq:closed-loop-gate}
\end{equation}
where $H$ is the number of completely correct tasks in the batch and $S$ is
mean cell accuracy. A failed edit is rolled back to the iteration-start
snapshot and retried from the same state, for at most three attempts.

\paragraph{No effectiveness signal in later open-loop diagnosis.}
Open-loop diagnosis receives neither a rejection context nor a prior record of
whether earlier edits changed accuracy. Its momentum record states what the
patcher changed, but the released open-loop artifacts contain no post-edit
outcome that a subsequent diagnosis could use. Closed-loop evolution adds two
channels:
\begin{itemize}
    \item \textbf{Within-iteration rejection context (F5):} after a rejected
    attempt, the next diagnosis receives the observed hard-score change and
    the failed edit direction, and is instructed not to repeat it.
    \item \textbf{Cross-iteration diagnosis prior (F6):} the most recent six
    accept/reject records, including hard-score changes and diagnosis labels,
    are injected into later attempts.
\end{itemize}

\subsubsection{Gate Trace for Seed~8}
\label{app:closed-loop-gate-trace}

The seed~8 closed-loop run contains ten iterations and 22 attempted edits.
Eight attempts reduce same-batch hard correctness and are blocked by the gate
(Table~\ref{tab:closed-loop-attempts}). In iteration~2, all three attempts
fail, so the entire iteration is reverted. By contrast, the ten patches
generated during open-loop evolution are all committed without post-update
verification.

\begin{apptable}
    \captionof{table}{Closed-loop attempt trace for seed~8. Each batch contains four
    tasks. An attempt whose post-update hard-correct count falls below the
    pre-update count is rejected.}
    \label{tab:closed-loop-attempts}
    \renewcommand{\arraystretch}{1.25}
    \setlength{\tabcolsep}{7pt}
    \arrayrulecolor{CaseTableBorder}
    \begin{tabularx}{\linewidth}{
        |>{\raggedleft\arraybackslash}p{0.07\linewidth}
        |>{\raggedleft\arraybackslash}p{0.07\linewidth}
        |>{\raggedright\arraybackslash}p{0.22\linewidth}
        |>{\raggedright\arraybackslash}X|}
        \hline
        \rowcolor{CaseTableHeader}
        \textbf{Iter.} & \textbf{Pre} &
        \textbf{Attempt post counts} & \textbf{Result} \\
        \hline
        1 & 2 & 1, 2, 2 &
        \textcolor{CasePositive}{\textbf{Accepted; attempt 1 regresses by 1 and is rejected.}}\\
        \hline
        2 & 3 & 2, 2, 1 &
        \textcolor{CaseNegative}{\textbf{All three attempts rejected; entire
        iteration reverted.}} \\
        \hline
        3 & 2 & 3 &
        \textcolor{CasePositive}{\textbf{Accepted; strict gain and early
        termination.}} \\
        \hline
        4 & 2 & 2, 2, 2 &
        \textcolor{CasePositive}{\textbf{Accepted; tie resolved using the best soft score.}} \\
        \hline
        5 & 1 & 1, 0, 1 &
        \textcolor{CasePositive}{\textbf{Accepted; attempt 2 regresses by 1 and is rejected.}} \\
        \hline
        6 & 0 & 0, 1 &
        \textcolor{CasePositive}{\textbf{Accepted; attempt 2 yields a strict
        gain.}} \\
        \hline
        7 & 3 & 3, 2, 2 &
        \textcolor{CasePositive}{\textbf{Accepted; attempts 2 and 3 regress by 1 and are rejected.}} \\
        \hline
        8 & 3 & 4 &
        \textcolor{CasePositive}{\textbf{Accepted; strict gain.}} \\
        \hline
        9 & 3 & 2, 4 &
        \textcolor{CasePositive}{\textbf{Attempt 1 rejected; attempt 2 yields
        a strict gain.}} \\
        \hline
        10 & 3 & 4 &
        \textcolor{CasePositive}{\textbf{Accepted; strict gain.}} \\
        \hline
    \end{tabularx}

    \vspace{3pt}
    \begin{minipage}{\linewidth}
        \footnotesize\color{gray}
        Across 10 iterations, the closed-loop procedure makes 22 attempts,
        blocks 8 hard regressions, and completely reverts one iteration.
    \end{minipage}
\end{apptable}
\arrayrulecolor{black}

This trace provides direct evidence that the patch--re-execute--gate mechanism
is active: regressive updates are intercepted online rather than permanently
written into the skill.

\subsubsection{Utility-Sign Flip on the Same Failure Pattern}
\label{app:closed-loop-utility-flip}

Both runs encounter a failure pattern involving sequential scans whose
reference value changes after each qualifying event. Both create a section for
this pattern, but the resulting knowledge differs substantially.

\begin{apptable}
    \captionof{table}{The same failure pattern produces qualitatively different
    knowledge in the matched open- and closed-loop runs.}
    \label{tab:closed-loop-formulation}
    \renewcommand{\arraystretch}{1.20}
    \setlength{\tabcolsep}{8pt}
    \begin{tabularx}{\linewidth}{
        >{\columncolor{red!3}\raggedright\arraybackslash}X
        >{\columncolor{green!4}\raggedright\arraybackslash}X}
        \toprule
        \textbf{Open-loop G1} & \textbf{Closed-loop G3f} \\
        \midrule
        \textbf{\emph{Trace Stateful Algorithms Before Coding}}
        &
        \textbf{\emph{Sequential Scan with Dynamic Reference}}
        \\[2pt]
        Requires the agent to manually trace at least one example step by
        step before coding.
        &
        States the operational semantics directly: maintain the reference as
        a variable that is updated during iteration.
        \\[2pt]
        Encodes a task-specific 19-line template with the threshold
        \texttt{1.10} hard-coded:\par
        \vspace{2pt}
        \texttt{count = 0}\par
        \texttt{reference = data[0]}\par
        \texttt{for i, val in enumerate(data[1:], start=1):}\par
        \texttt{\phantom{xx}if val >= reference * 1.10:}\par
        \texttt{\phantom{xxxx}count += 1}\par
        \texttt{\phantom{xxxx}if i + 1 < len(data):}\par
        \texttt{\phantom{xxxxxx}reference = data[i + 1]}
        &
        Uses a short parameterized pattern:\par
        \vspace{2pt}
        \texttt{reference = first\_value}\par
        \texttt{count = 0}\par
        \texttt{for value in remaining\_values:}\par
        \texttt{\phantom{xx}if value >= reference * threshold:}\par
        \texttt{\phantom{xxxx}count += 1}\par
        \texttt{\phantom{xxxx}reference = value}
        \\[2pt]
        Its final rule is a difficult-to-falsify meta-instruction: if a small
        example cannot be manually verified, the logic is not ready to code.
        &
        Its final decision rule is executable: update the reference to the
        current value after every qualifying event; do not compare all values
        against one static baseline.
        \\[2pt]
        \textcolor{CaseNegative}{\textbf{Negative: }
        $u^{\mathrm{cell}}=-0.0337$,
        $u^{\mathrm{hard}}=-0.0556$}
        &
        \textcolor{CasePositive}{\textbf{Positive and protected: }
        $u^{\mathrm{cell}}=+0.0495$,
        $u^{\mathrm{hard}}=+0.0474$}
        \\
        \bottomrule
    \end{tabularx}
\end{apptable}

As shown in Table~\ref{tab:closed-loop-formulation}, the cell utility changes
from $-0.0337$ to $+0.0495$, a difference of $0.0832$.

The difference is actionability rather than length. Advice such as ``trace
carefully'' need not produce a measurable behavioral change on same-batch
re-execution. By contrast, updating the reference after each event changes the
generated program and can therefore be tested by the gate. A rejected edit
also supplies the next attempt with evidence that its direction was
ineffective. In this sense, the gate selects for \emph{falsifiable edits}:
their behavioral consequences can be observed immediately, accepted when
useful, and rejected when regressive. This mechanism is consistent with the
four repeated variants of the ``trace an example'' meta-instruction found in
the open-loop skill (Section~\ref{app:prox-case-redundancy}).

\subsubsection{What Closed-Loop Evolution Does Not Solve}
\label{app:closed-loop-residual}

Closed-loop evolution is not a complete negative-utility filter. The
leave-one-out audits are summarized in
Table~\ref{tab:closed-loop-audit-summary}.

\begin{apptable}
    \captionof{table}{Utility composition and main-file length for the matched seed~8
    skills. Lengths use the verified final \texttt{SKILL.md} artifacts.}
    \label{tab:closed-loop-audit-summary}
    \setlength{\tabcolsep}{6pt}
    \begin{tabular}{lrr}
        \toprule
        & Open-loop G1 & Closed-loop G3f \\
        \midrule
        Knowledge units & 16 & 22 \\
        Positive & 4 (25.0\%) & 6 (27.3\%) \\
        Zero & 10 & 14 \\
        Negative & 2 (12.5\%) & 2 (9.1\%) \\
        \texttt{SKILL.md} characters & 21{,}191 & 23{,}748 \\
        \bottomrule
    \end{tabular}
\end{apptable}

Two observations matter. First, the closed-loop skill is longer, not shorter
(+2{,}557 characters, +12.1\%); the mechanism performs quality control rather
than compression. Second, it still contains two estimated-negative units. One
is the reference file \texttt{sequential\_scan.md}
($u^{\mathrm{cell}}=-0.0395$) attached to the positive-utility main section
analyzed above. The batch-level gate can therefore miss content that is
harmless on its originating training batch but harmful under broader
validation. Backward Prox remains complementary
(Section~\ref{app:prox-case-study}): on this seed, G1 scores 46 OJ hard,
G3f scores 51, and G3f followed by Prox-D scores 54 under the corresponding
canonical evaluations.

\subsection{Case Study: Consolidating Negative-Utility Redundancy}
\label{app:prox-case-study}

\subsubsection{Negative-Utility Redundancy}
\label{app:prox-case-redundancy}

The open-loop skill for seed~8 expresses the meta-instruction ``trace a
concrete example before coding'' in four separate sections.
Leave-one-out auditing on the 20-task validation set with
$\epsilon_{\mathrm{cell}}=0.025$ assigns these sections substantially
different cell utilities (Table~\ref{tab:prox-case-repeated-instructions}).

\begin{apptable}
    \captionof{table}{Four occurrences of the same meta-instruction in the evolved
    open-loop skill. Utilities are estimated by leave-one-out evaluation on
    20 validation tasks.}
    \label{tab:prox-case-repeated-instructions}
    \setlength{\tabcolsep}{3pt}
    \begin{tabular}{p{0.42\linewidth}r}
        \toprule
        Section & $u^{\mathrm{cell}}$ \\
        \midrule
        Ground Task Interpretation
            & $+0.1038$ \\
        Interpreting Aggregation Language
            & $+0.0119$ \\
        Verify Ambiguous Arithmetic Direction
            & $-0.0212$ \\
        Trace Stateful Algorithms Before Coding
            & $-0.0337$ \\
        \bottomrule
    \end{tabular}
\end{apptable}

The lowest-utility section, \emph{Trace Stateful Algorithms Before Coding},
is the same negative-utility formulation identified in
Section~\ref{app:closed-loop-utility-flip}
(Table~\ref{tab:closed-loop-formulation}).
It does more than repeat the general advice: it presents a
training-task-specific implementation as a reusable template, including the
hard-coded condition \texttt{value >= reference * 1.10}.
Thus the section is estimated to be harmful on the validation audit, rather
than merely dispensable. Because the audit uses only 20 validation tasks, we
treat these scores as candidate evidence rather than as definitive causal
effects.

\subsubsection{Frozen Audit versus Dynamic Validation Gate}
\label{app:prox-case-gate}

The audit evaluates 16 knowledge units and classifies two as negative, ten as
zero-utility, and four as positive. Prox considers units with
$u^{\mathrm{cell}}<-0.001$ as shrinkage candidates. Each realized removal is
then re-evaluated against the current active skill under
$\delta_h=0$, $\delta_c=0.02$, and a soft compression cap $\rho=0.10$. Any
trial that violates the validation gate is rolled back.

\begin{apptable}
    \captionof{table}{Candidate-level audit scores and realized validation decisions.
    A negative audit score determines eligibility but does not by itself
    authorize deletion.}
    \label{tab:prox-case-decisions}
    \setlength{\tabcolsep}{2.5pt}
    \begin{tabular}{lrrp{0.28\linewidth}}
        \toprule
        Candidate
        & $u^{\mathrm{cell}}$
        & $u^{\mathrm{hard}}$
        & Decision \\
        \midrule
        Trace Stateful Algorithms
            & $-0.0337$ & $-0.0556$
            & Accepted \\
        Error Handling in Aggregation
            & $-0.0337$ & $-0.0556$
            & Rolled back \\
        Verify Ambiguous Arithmetic
            & $-0.0212$ & $-0.0056$
            & Rolled back \\
        Reading Data with pandas
            & $-0.0067$ & $-0.0030$
            & Rolled back \\
        Inspect Before Operating
            & $-0.0017$ & $-0.0030$
            & Rolled back \\
        \bottomrule
    \end{tabular}
\end{apptable}

\begin{appfigure}
    \begin{tikzpicture}
        \begin{axis}[
            width=0.6\linewidth,
            height=5.2cm,
            ybar,
            ymin=0,
            ymax=100,
            ylabel={Accuracy (\%)},
            symbolic x coords={Hard,Cell},
            xtick=data,
            xticklabels={{Hard / perfect tasks},{Mean cell accuracy}},
            xticklabel style={align=center,font=\scriptsize},
            ylabel style={font=\small},
            yticklabel style={font=\scriptsize},
            bar width=16pt,
            enlarge x limits=0.35,
            nodes near coords,
            nodes near coords style={font=\scriptsize},
            legend style={
                at={(0.5,-0.28)},
                anchor=north,
                legend columns=2,
                draw=none,
                font=\scriptsize
            },
            grid=major,
            grid style={draw=black!8},
            axis line style={draw=black!45},
        ]
            \addplot[fill=ProxBefore,draw=none]
                coordinates {(Hard,46.00) (Cell,74.71)};
            \addplot[fill=ProxAfter,draw=none]
                coordinates {(Hard,54.00) (Cell,77.97)};
            \legend{Before (G1),After Prox}
        \end{axis}
    \end{tikzpicture}
    \captionof{figure}{OJ accuracy before and after validation-gated Prox consolidation.
    Evaluation uses 100 fixed tasks with three test cases per task; hard
    success requires all three cases to be cell-perfect. Values are not
    smoothed.}
    \label{fig:prox-case-oj}
\end{appfigure}

Five candidates are attempted, but only one edit is accepted
(Table~\ref{tab:prox-case-decisions}). The four rejected edits are important:
although their frozen audit scores are negative or near-zero, deleting them
from the current active skill reduces validation hard or cell accuracy.
For example, removing \emph{Verify Ambiguous Arithmetic Direction} lowers hard
accuracy by 5.85 points. The frozen audit therefore identifies and orders
candidates; an independent validation gate remains necessary before an edit is
committed.

\subsubsection{Structural Consolidation and OJ Outcome}
\label{app:prox-case-oj}

The accepted edit is a consolidation rather than a pure deletion. Prox removes
the task-specific template but retains the transferable principle as a concise
sentence in the positive-utility \emph{Ground Task Interpretation in Data
Patterns} section. All high-positive-utility units remain unchanged.

\begin{apptable}
    \captionof{table}{Structural and OJ changes produced by the accepted Prox edit.
    OJ hard accuracy uses the fixed 100-task denominator.}
    \label{tab:prox-case-summary}
    \setlength{\tabcolsep}{4pt}
    \begin{tabular}{lrrr}
        \toprule
        Measure & Before & After & Change \\
        \midrule
        Complete skill characters
            & 29{,}129 & 28{,}219 & $-3.12\%$ \\
        \texttt{SKILL.md} characters
            & 21{,}191 & 20{,}281 & $-4.29\%$ \\
        Second-level sections
            & 14 & 13 & $-1$ \\
        Validation hard
            & $94.74\%$ & $94.74\%$ & $0$ \\
        Validation cell
            & $96.05\%$ & $99.73\%$ & $+3.68$ \\
        OJ hard
            & $46\%$ & $54\%$ & $+8$ \\
        OJ mean cell
            & $74.71\%$ & $77.97\%$ & $+3.26$ \\
        Fail-to-pass / pass-to-fail
            & \multicolumn{2}{c}{$8$ / $0$} & \\
        \bottomrule
    \end{tabular}
\end{apptable}

Table~\ref{tab:prox-case-summary} summarizes the outcome. The complete skill
shrinks from 29{,}129 to 28{,}219 characters,

\subsection{Discussion}
\label{app:case-discussion}

Validation hard accuracy remains at $94.74\%$, while validation cell accuracy
increases from $96.05\%$ to $99.73\%$. On the independent OJ evaluation, hard
accuracy increases from $46\%$ to $54\%$ and mean cell accuracy from $74.71\%$
to $77.97\%$ (Figure~\ref{fig:prox-case-oj}). The task-level comparison contains eight fail-to-pass transitions
and no pass-to-fail transition; one of the eight transitions is caused by an
inference API error in the pre-Prox condition and should not be interpreted as
a skill effect. Conservatively excluding this task leaves seven graded
fail-to-pass improvements (Table~\ref{tab:prox-case-transitions}).

\begin{apptable}
    \captionof{table}{Task-level fail-to-pass transitions after Prox consolidation.
    No task exhibits a pass-to-fail transition.}
    \label{tab:prox-case-transitions}
    \renewcommand{\arraystretch}{1.18}
    \setlength{\tabcolsep}{7pt}
    \begin{tabularx}{\linewidth}{
        >{\raggedright\arraybackslash}p{0.10\linewidth}
        >{\raggedright\arraybackslash}p{0.08\linewidth}
        >{\raggedleft\arraybackslash}p{0.16\linewidth}
        >{\raggedright\arraybackslash}X}
        \toprule
        Task & Type & Cell accuracy & Behavioral difference \\
        \midrule
        469-9 & Sheet & $0.100\rightarrow1.000$ &
        Correctly determines the debit/credit sign direction. \\
        230-16 & Sheet & $0.222\rightarrow1.000$ &
        Scans upward to find the true final row instead of trusting
        \texttt{max\_row}. \\
        38537 & Cell & $0.441\rightarrow1.000$ &
        Identifies the true data range \texttt{B3:B36}. \\
        35739 & Cell & $0.475\rightarrow1.000$ &
        Correctly handles a cutoff time crossing midnight. \\
        38969 & Cell & $0.526\rightarrow1.000$ &
        Corrects the nested \texttt{IF} logic and column scan range. \\
        51090 & Cell & $0.818\rightarrow1.000$ &
        Completes multi-sheet filtered aggregation ($18/22\rightarrow22/22$). \\
        16511 & Cell & $0.889\rightarrow1.000$ &
        Assigns increasing track numbers within each album. \\
        \bottomrule
    \end{tabularx}
\end{apptable}

The graded improvements concentrate on data-range identification and task
semantics, which are precisely the roles of the retained inspection and
interpretation sections. This correspondence is descriptive: because the two
OJ conditions are independent generations, it does not identify the accepted
edit as the unique cause of each transition.
These OJ results should be interpreted as an associated improvement under an
independent generation at temperature $0.7$, rather than as a strictly causal
effect of the single accepted edit. The stronger causal claim supported by this
case is local: validation-gated Prox identifies a negative-utility formulation,
removes its task-specific template, and consolidates the transferable principle
without degrading the validation gate.

The two case studies examine complementary treatments of related accumulation
failures on the same seed. Closed-loop forward evolution provides online
interception: it measures the immediate batch-level effect of a proposed update
and blocks hard regressions. Backward Prox provides retrospective refinement:
it reassesses the accumulated skill on a frozen validation set and consolidates
redundant or estimated-negative content that was not detectable from the
originating batch. Applying Prox-D to the closed-loop seed~8 skill yields an OJ
hard accuracy of $54\%$, consistent with the view that the two stages compose.

These findings should be interpreted cautiously. First, both case studies
reuse seed~8, which was selected because it exhibits the largest closed-loop
gain and is also the weakest open-loop run; its $+5$ gain must not be
extrapolated as an average treatment effect. Second, the two skill trajectories
diverge early, so the utility-sign flip is observational rather than a
controlled A/B edit. Third, the OJ conditions are independent generations at
temperature $0.7$, not paired rerolls with fixed sampling randomness. Fourth,
the utility audit uses only 20 validation tasks. Finally, the released
seed~8 closed-loop results do not record \texttt{pre\_soft} or attempt-level
rejection labels; the attempt-level hard-regression annotations in
Table~\ref{tab:closed-loop-attempts} are reconstructed by comparing the
recorded pre- and post-correct counts, while iteration-level labels are
available. The value of the cases is
therefore mechanistic: they show how online outcome gating and post-training
Prox address different timescales of the same accumulation problem.

\subsection{Additional Analysis of Model Size and Skill Updates}
\label{app:model-size}

\subsubsection{Final Skill Length and IID Accuracy}
We examine whether a longer final skill is associated with stronger in-domain
performance. Here, the final skill refers to G3D, i.e., the skill obtained
after closed-loop training and Prox-D. Skill length is measured as the total
number of Markdown characters in the main \texttt{SKILL.md} file and its
reference files; agent-generated scratch files are excluded. IID performance
is measured using SpreadsheetBench OJ hard accuracy over the fixed 100-task
test set, where one generated program is applied to all three test cases and a
task is counted as correct only if all three cases pass.
Figure~\ref{fig:g3d-length-iid} shows the nine Qwen3.5-27B training seeds.
Final skill length is negatively associated with IID hard accuracy
($r=-0.628$, $n=9$). In particular, the longest skills do not achieve the
highest accuracy, suggesting that accumulating more textual guidance is not
necessarily beneficial. The fitted line is intended as a descriptive summary,
not a causal relationship. The estimate is also sensitive to seed~8; removing
this seed reduces the correlation to $r=-0.330$. Therefore, the result should
be interpreted as evidence that skill length alone is not a reliable proxy for
skill quality, rather than as a statistically conclusive monotonic effect.

\begin{appfigure}
    \includegraphics[width=0.6\linewidth]{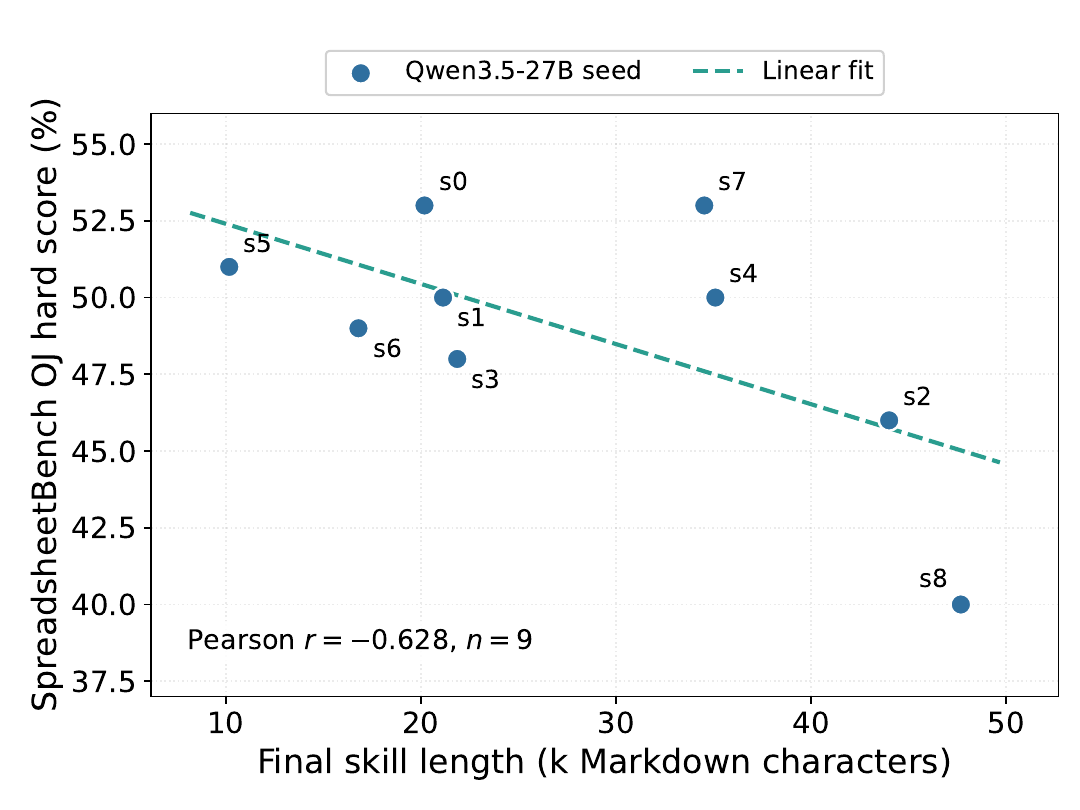}
    \captionof{figure}{Final G3D skill length versus SpreadsheetBench IID OJ hard
    accuracy for nine Qwen3.5-27B training seeds. Length includes Markdown
    content in the main skill file and reference files. The dashed line is an
    ordinary least-squares fit.}
    \label{fig:g3d-length-iid}
\end{appfigure}

\subsubsection{Closed-Loop Update Dynamics}
To understand why model size affects the final skill representation, we
analyze the inner closed-loop trajectories (Table~\ref{tab:closed-loop-dynamics}). Since one
diagnosis may trigger multiple writes, writes per diagnosis is an update
intensity proxy rather than a literal probability.

\begin{apptable}
    \captionof{table}{Closed-loop update dynamics. Values are mean $\pm$ sample
    standard deviation over three 4B runs and nine 27B runs.}
    \label{tab:closed-loop-dynamics}
    \setlength{\tabcolsep}{4pt}
    \begin{tabular}{lcc}
        \toprule
        Metric & Qwen3.5-4B & Qwen3.5-27B \\
        \midrule
        Attempts / iteration
            & $2.57\pm0.15$ & $2.13\pm0.36$ \\
        Diagnoses / run
            & $102.7\pm6.1$ & $85.3\pm14.6$ \\
        Characters / diagnosis
            & $1355\pm75$ & $1474\pm129$ \\
        Write calls / run
            & $57.3\pm9.9$ & $35.6\pm8.2$ \\
        Accepted iterations
            & $66.7\%\pm15.3\%$ & $85.6\%\pm10.1\%$ \\
        G1 writes / diagnosis
            & $0.96$ & $0.40$ \\
        G3f writes / diagnosis
            & $0.56$ & $0.42$ \\
        Realized Prox compression
            & $29.4\%\pm17.9\%$ & $19.0\%\pm18.1\%$ \\
        \bottomrule
    \end{tabular}
\end{apptable}

The 4B model requires more inner-loop attempts, generates more diagnoses, and
performs substantially more write calls than the 27B model. More importantly,
the number of writes per diagnosis decreases from 0.96 under open-loop G1
training to 0.56 under closed-loop G3f training for the 4B model. In contrast,
the corresponding value for the 27B model remains nearly unchanged
($0.40\rightarrow0.42$). The gate also rejects more final updates from the 4B
model: only $66.7\%$ of its iterations retain the proposed update, compared
with $85.6\%$ for 27B.
These observations suggest that the larger model is already relatively
selective about whether a diagnosis should modify the skill, whereas the
smaller model tends to translate diagnoses into edits more aggressively under
open-loop training. The patch--re-execute--gate mechanism therefore acts as an
external update filter that disproportionately regularizes the smaller model.
Prox-D reinforces the same effect by applying stronger realized compression to
the 4B skills than to the 27B skills ($29.4\%$ versus $19.0\%$). Together,
closed-loop validation and proximal shrinkage move the smaller model toward a
more selective and compact update regime.

\clearpage
\section{Prompts and Algorithm}\label{app:prompt_templates}
\input{prompts_and_algorithm}

%% file: prompts_and_algorithm.tex

%

\definecolor{ExecutorBlue}{RGB}{128,119,239}
\definecolor{FailureRed}{RGB}{245,139,143}
\definecolor{ContrastiveOrange}{RGB}{231,152,86}
\definecolor{FeedbackPurple}{RGB}{157,116,214}
\definecolor{MomentumTeal}{RGB}{62,157,153}
\definecolor{PatcherGreen}{RGB}{75,154,103}
\definecolor{ShrinkerGold}{RGB}{194,145,52}

\newtcblisting{promptbox}[2]{
    enhanced,
    breakable,
    listing only,
    colback=#2!5,
    colframe=#2!65,
    colbacktitle=#2,
    coltitle=white,
    fonttitle=\bfseries,
    title={#1},
    boxrule=0.6pt,
    arc=1.5pt,
    outer arc=1.5pt,
    left=2.5mm,
    right=2.5mm,
    top=1.5mm,
    bottom=1.5mm,
    listing options={
        basicstyle=\ttfamily\footnotesize,
        breaklines=true,
        breakatwhitespace=true,
        columns=fullflexible,
        keepspaces=true,
        showstringspaces=false,
        upquote=true
    }
}

\newcounter{skillproxalgorithm}
\renewcommand{\theskillproxalgorithm}{\arabic{skillproxalgorithm}}
\providecommand{\Call}[2]{\operatorname{#1}\!\left(#2\right)}

\newenvironment{algorithmBox}[2][]{%
    \refstepcounter{skillproxalgorithm}%
    \begin{tcolorbox}[
        enhanced,
        breakable,
        colback=black!1,
        colframe=black!55,
        colbacktitle=black!8,
        coltitle=black,
        fonttitle=\bfseries,
        title={Algorithm~\theskillproxalgorithm: #2},
        boxrule=0.6pt,
        arc=1.5pt,
        outer arc=1.5pt,
        left=1.5mm,
        right=1.5mm,
        top=1mm,
        bottom=1mm,
        #1
    ]%
}{%
    \end{tcolorbox}%
}

\label{app:prompts}

This appendix gives some representative prompts in SkillProx, Some other prompts follow the setting of SkillGrad~\cite{wang2026skillgrad}.
In the end, we provide the overall algorithm in SkillProx.

\subsection{Momentum Prompt}
\label{app:prompt-momentum}

The momentum agent converts per-task diagnoses into a persistent pattern
record and a current-iteration overlay. The former carries recurrence and
remedy history; the latter gives the patcher one actionable entry per task.

\begin{promptbox}{Pattern-record writer (momentum) prompt}{MomentumTeal}
[SYSTEM]
You are PatternRecordWriter.

Read:
1. batch_diagnoses.md;
2. the previous momentum_memory.md (empty at iteration 1); and
3. the current SKILL.md and references.

Write:
1. an updated cross-iteration pattern record; and
2. a per-task overlay in the same order as the diagnoses.

A pattern is a class of mistake or success, not a task instance. Default to
merging evidence into an existing pattern. Split only when the required
operations or corrective actions genuinely differ. Preserve remedy_log as
append-only history.

Pattern record entry:
### {pattern-id} | {operation|workflow|mixed} | {description}
- anchor: {L2 section slug | L3 filename | (none yet)}
- appeared_in: {iterations}
- description: {accumulated evidence}
- latest_executor_action: {current actionable rule}
- remedy_log:
  - iter_{k} | diagnosis: {summary} | patch: {action}

Overlay entry:
### [{task_id}] {signal description}
- signal: failure | success
- pattern: {pattern-id | new | no-actionable-signal}
- anchor: {anchor}
- gap: {missing or misfiring behavior}
- proposed_change: {targeted edit direction}

If the diagnosis is too vague, use "no-actionable-signal" and omit gap and
proposed_change. End with at most three recurrent WORKFLOW-THEMES.

[USER]
This is iteration {k}.

Inputs:
  Current iteration diagnoses: {batch_diagnoses_path}
  Previous pattern record: {previous_memory_path}
  Current guidance files:
    - {skill_md_path}
    - {reference_paths}

Outputs:
  Updated pattern record: {memory_output_path}
  Per-task overlay: {overlay_output_path}
\end{promptbox}

\subsection{Patcher Prompt}
\label{app:prompt-patcher}

The patcher applies one layer-aware update to the skill. It reads both the
raw diagnoses and the momentum outputs so that compression by the momentum
agent cannot erase task-level evidence.

\begin{promptbox}{Skill patcher prompt}{PatcherGreen}
[SYSTEM]
You are Patcher. Edit the existing skill in place; never redraft it wholesale.

The skill has three layers:
- L1 metadata routes skill selection.
- L2 SKILL.md is always loaded and must remain general and skimmable.
- L3 references are loaded conditionally and hold specific operations,
  runtime branches, worked examples, and verification procedures.

Iterate by pattern, not by task:
1. Group overlay entries sharing a pattern.
2. Read their pattern history and raw diagnoses.
3. Brainstorm 2--3 candidate remedies.
4. Apply the simplest edit that generalizes.

Use YAGNI. Prefer extending or merging an existing section over creating a
new one. Never place task-specific columns, rows, filenames, or constants in
L2. Every L3 file must have exactly one L2 pointer and must contain runnable
code, a runtime branch, and a verification step with a corrective action.

After editing, read back all changed files and repair broken pointers,
orphaned references, duplicate sections, and workflow-checklist growth.

[USER]
Evolve the skill based on the analysis below.

Original diagnoses: {batch_diagnoses_path}
Per-attempt overlay: {momentum_overlay_path}
Cross-iteration pattern record: {momentum_memory_path}

Skill files:
  - {skill_md_path}
  - {reference_paths}

Reference directory for new L3 files: {references_directory}
\end{promptbox}

\subsection{Shrinker Prompt}
\label{app:prompt-shrinker}

The utility audit itself has no generative prompt: it repeatedly calls the
executor on the full and leave-out skills. Once the deterministic audit
selects a candidate, the Shrinker receives exactly one target unit.

\begin{promptbox}{Prox shrinker prompt}{ShrinkerGold}
[SYSTEM]
You are Shrinker. The forward optimizer only added content; make the skill
smaller and cleaner without losing what works.

The target is an L2 section or L3 reference whose measured marginal utility
is low or negative. Diagnose it as:
(a) redundant, (b) contradictory or misleading,
(c) over-specific or verbose, or (d) genuinely load-bearing.

Apply the lightest valid operation:
1. Consolidate (preferred): salvage unique generalizable content into the
   most-overlapping retained section, then remove the target.
2. Demote: move excessive L2 detail into a concise L3 procedure.
3. Remove: delete content that is fully redundant or misleading.

Touch only the target and, when needed, one receiving section. Preserve all
L2--L3 pointer invariants. A downstream validation gate will reject the edit
if accuracy drops. The resulting skill must be strictly smaller.

[USER]
Shrink the skill by processing exactly one target unit.

Target unit (flagged low/negative marginal value):
{target_L2_section_or_L3_reference}

Skill files:
  - {skill_md_path}
  - {reference_paths}

Salvage genuinely useful, generalizable content from the target into the
most-overlapping retained section. Remove the redundant, contradictory, or
verbose remainder. Leave all other sections unchanged.
\end{promptbox}

\subsection{End-to-End SkillProx}
\label{app:full-pseudocode}

Algorithm~\ref{alg:skillprox} covers data construction, open- or
closed-loop forward evolution, leave-one-out utility estimation, and the
validation-gated Prox backward pass. The closed-loop branch describes the
multi-attempt behavior represented by the released run artifacts: rejected
edits are rolled back and retried up to three times, with feedback
injected into the next diagnosis.

\begin{algorithmBox}
{SkillProx: closed-loop forward evolution and Prox consolidation}
\label{alg:skillprox}
\footnotesize
\begin{algorithmic}[1]
\REQUIRE Initial skill $S_0$; model $M$; evolution pool $\mathcal{D}_e$;
validation set $\mathcal{V}$; mode
$m\in\{\mathrm{open},\mathrm{closed}\}$; batch size $b$; iterations $K$;
attempts $A$; utility threshold $\tau$; compression cap $\rho$; validation
tolerances $(\delta_h,\delta_c)$
\ENSURE Forward skill $S_{\mathrm{fwd}}$ and consolidated skill
$S_{\mathrm{prox}}$

\STATE \textbf{Phase I: Construct failure-driven update batches}
\STATE $R_0\gets\Call{ExecuteAndAssess}{M,S_0,\mathcal{D}_e}$
\STATE $\mathcal{F}\gets
\{x\in\mathcal{D}_e:\Call{Cell}{R_0[x]}<1\}$
\STATE $\mathcal{T}\gets
\Call{SampleUpTo40}{\mathcal{F},\mathrm{training\ seed}}$
\STATE $(B_1,\ldots,B_K)\gets\Call{Chunk}{\mathcal{T},b}$
\STATE $S\gets\Call{Copy}{S_0}$; $P\gets\emptyset$;
$Q\gets\emptyset$; $c_{\mathrm{all}}\gets0$

\STATE \textbf{Phase II: Forward skill evolution}
\FOR{$k=1,\ldots,K$}
    \STATE $S^{(0)}\gets\Call{Copy}{S}$ \COMMENT{iteration snapshot}
    \STATE $R_{\mathrm{pre}}\gets\Call{ExecuteAndAssess}{M,S,B_k}$
    \STATE $h_{\mathrm{pre}}\gets\Call{Hard}{R_{\mathrm{pre}}}$;
    $s_{\mathrm{pre}}\gets\Call{Cell}{R_{\mathrm{pre}}}$
    \STATE $(\mathcal{F}_k,\mathcal{C}_k)\gets
    \Call{Classify}{R_{\mathrm{pre}}}$

    \IF{$\mathcal{F}_k=\emptyset$}
        \STATE $c_{\mathrm{all}}\gets c_{\mathrm{all}}+1$
        \IF{$c_{\mathrm{all}}=4$}
            \STATE \textbf{break}
        \ENDIF
    \ELSE
        \STATE $c_{\mathrm{all}}\gets0$
    \ENDIF

    \STATE $a_{\max}\gets A$ if $m=\mathrm{closed}$, else $1$
    \STATE $C\gets\emptyset$ \COMMENT{gate-eligible candidate skills}
    \STATE $r\gets\emptyset$ \COMMENT{within-iteration rejection context}

    \FOR{$a=1,\ldots,a_{\max}$}
        \IF{$a>1$}
            \STATE $S\gets\Call{Restore}{S^{(0)}}$
        \ENDIF
        \STATE $q\gets\Call{FormatPrior}{Q,\mathrm{last}=6}$
        \STATE $D_k\gets
        \Call{Diagnose}{M,\mathcal{F}_k,\mathcal{C}_k,
        R_0,R_{\mathrm{pre}},q,r}$

        \IF{$k=1$}
            \STATE $O_k\gets\emptyset$
            \COMMENT{bootstrap has no prior momentum}
        \ELSE
            \STATE $(P,O_k)\gets\Call{Momentum}{M,D_k,P,S}$
        \ENDIF

        \STATE $\widetilde S\gets\Call{Patch}{M,S,D_k,P,O_k}$
        \IF{$m=\mathrm{open}$}
            \STATE $S\gets\widetilde S$
            \COMMENT{open-loop patch is committed}
            \STATE \textbf{break}
        \ELSE
            \STATE $R_{\mathrm{post}}\gets
            \Call{ExecuteAndAssess}{M,\widetilde S,B_k}$
            \STATE $h_{\mathrm{post}}\gets\Call{Hard}{R_{\mathrm{post}}}$;
            $s_{\mathrm{post}}\gets\Call{Cell}{R_{\mathrm{post}}}$
            \STATE $g\gets
            [h_{\mathrm{post}}\ge h_{\mathrm{pre}}]\land
            [s_{\mathrm{post}}\ge s_{\mathrm{pre}}-10^{-9}]$

            \IF{$g$}
                \STATE $C\gets C\cup
                \{(h_{\mathrm{post}},s_{\mathrm{post}},
                \Call{Copy}{\widetilde S},D_k)\}$
                \IF{$h_{\mathrm{post}}>h_{\mathrm{pre}}$}
                    \STATE \textbf{break}
                    \COMMENT{strict hard improvement}
                \ENDIF
            \ELSE
                \STATE $S\gets\Call{Restore}{S^{(0)}}$
                \STATE $r\gets\Call{RejectContext}{
                h_{\mathrm{pre}},s_{\mathrm{pre}},
                h_{\mathrm{post}},s_{\mathrm{post}},\widetilde S}$
            \ENDIF
        \ENDIF
    \ENDFOR

    \IF{$m=\mathrm{closed}$}
        \IF{$C\neq\emptyset$}
            \STATE $(h^*,s^*,S,D_k)\gets
            \operatorname*{arg\,max}_{(h,s,\cdot,\cdot)\in C}(h,s)$
            \STATE $Q\gets Q\cup
            \{(k,\mathrm{accepted},h_{\mathrm{pre}},h^*)\}$
        \ELSE
            \STATE $S\gets\Call{Restore}{S^{(0)}}$
            \STATE $Q\gets Q\cup
            \{(k,\mathrm{rejected},h_{\mathrm{pre}},h_{\mathrm{pre}})\}$
        \ENDIF
    \ENDIF

    \IF{$k=1$}
        \STATE $(P,\_)\gets\Call{Momentum}{M,D_k,\emptyset,S}$
        \COMMENT{bootstrap persistent pattern record}
    \ENDIF
\ENDFOR
\STATE $S_{\mathrm{fwd}}\gets\Call{Copy}{S}$

\STATE \textbf{Phase III: Leave-one-out utility audit}
\STATE $U\gets\Call{ParseUnits}{S_{\mathrm{fwd}}}$
\STATE $v_0\gets\Call{Evaluate}{M,S_{\mathrm{fwd}},\mathcal{V}}$
\FORALL{$u_i\in U$}
    \STATE $S_{-i}\gets\Call{MaterializeWithout}{S_{\mathrm{fwd}},u_i}$
    \STATE $v_{-i}\gets\Call{Evaluate}{M,S_{-i},\mathcal{V}}$
    \STATE $u_i^{\mathrm{cell}}\gets
    v_0^{\mathrm{cell}}-v_{-i}^{\mathrm{cell}}$
    \STATE $u_i^{\mathrm{hard}}\gets
    v_0^{\mathrm{hard}}-v_{-i}^{\mathrm{hard}}$
\ENDFOR
\STATE $L\gets\Call{SortAscending}{
\{u_i:u_i^{\mathrm{cell}}<\tau\},
(u_i^{\mathrm{cell}},u_i^{\mathrm{hard}})}$

\STATE \textbf{Phase IV: Validation-gated Prox consolidation}
\STATE $S_{\mathrm{prox}}\gets\Call{Copy}{S_{\mathrm{fwd}}}$
\STATE $v\gets\Call{Evaluate}{M,S_{\mathrm{prox}},\mathcal{V}}$;
$n_0\gets|S_{\mathrm{prox}}|$
\FORALL{$u_i\in L$}
    \IF{$(n_0-|S_{\mathrm{prox}}|)/n_0\ge\rho$}
        \STATE \textbf{break}
    \ENDIF
    \STATE $T\gets\Call{Copy}{S_{\mathrm{prox}}}$
    \STATE $T\gets\Call{Shrink}{M,T,u_i}$
    \COMMENT{consolidate, demote, or remove}
    \STATE $T\gets\Call{GarbageCollectOrphanReferences}{T}$
    \IF{$\neg\Call{StructurallyValid}{T}\lor
          |T|\ge|S_{\mathrm{prox}}|$}
        \STATE \textbf{continue}
    \ENDIF
    \STATE $\widetilde v\gets\Call{Evaluate}{M,T,\mathcal{V}}$
    \IF{$\widetilde v^{\mathrm{hard}}\ge
        v^{\mathrm{hard}}-\delta_h$ \textbf{and}
        $\widetilde v^{\mathrm{cell}}\ge
        v^{\mathrm{cell}}-\delta_c$}
        \STATE $S_{\mathrm{prox}}\gets T$;
        $v\gets\widetilde v$
        \COMMENT{otherwise the trial is discarded}
    \ENDIF
\ENDFOR
\RETURN $S_{\mathrm{fwd}},S_{\mathrm{prox}}$
\end{algorithmic}
\end{algorithmBox}